\documentclass[journal]{IEEEtran}

\usepackage[dvipsnames]{xcolor}
\usepackage{graphicx}
\usepackage{amsmath}
\usepackage{amssymb}
\usepackage{booktabs}
\usepackage{multirow}
\usepackage{cite}
\usepackage[hyphens]{url}

\newcommand{\auspicesstandard}{This work was performed under the auspices of the
  U.S.\ Department of Energy by Lawrence Livermore National Laboratory under
  Contract DE-AC52-07NA27344.}

\newcommand{\papertitle}{Variable-Granularity Tokenization for High-Resolution Object Detection}

\usepackage{etoolbox}
\AtEndPreamble{%
  \usepackage[capitalize]{cleveref}%
  \crefname{section}{Sec.}{Secs.}%
  \Crefname{section}{Section}{Sections}%
  \Crefname{table}{Table}{Tables}%
  \crefname{table}{Tab.}{Tabs.}%
  \crefname{appendix}{App.}{Apps.}%
  \Crefname{appendix}{Appendix}{Appendices}%
}

\makeatletter
\def\@IEEEsectpunct{\ }
\makeatother

\AtBeginEnvironment{abstract}{\boldmath}

\usepackage{float}
\makeatletter
\newcommand{\floatsintextorder}{%
  \raggedbottom
  \crefalias{section}{appendix}%
  \let\vgtok@xfloat\@xfloat
  \def\@xfloat##1[##2]{\vgtok@xfloat{##1}[H]}%
  \@namedef{endtable*}{\endtable}%
  \@namedef{endfigure*}{\endfigure}%
  \patchcmd{\figure}{\@IEEEfiguretopskipspace}%
    {\@IEEEfiguretopskipspace\columnwidth=\dimexpr(\textwidth-\columnsep)/2\relax}%
    {}{\PackageError{preamble_ieee}{IEEEtran \string\figure\space changed}{}}}
\makeatother
\usepackage[hidelinks]{hyperref}

\title{\papertitle}

\author{Khayrul~Islam\thanks{This work has been submitted to the IEEE for possible publication. Copyright may be transferred without notice, after which this version may no longer be accessible.}\thanks{\auspicesstandard\ Release number: LLNL-CONF-2023636.}\thanks{K. Islam is with Lawrence Livermore National Laboratory, Livermore, CA 94550 USA (e-mail: islam11@llnl.gov).}}

\begin{document}
\bstctlcite{IEEEexample:BSTcontrol}
\maketitle

\begin{abstract}
ViT detectors fix a uniform token grid before any learned stage. A native-resolution aerial detector must then choose between resolving few-pixel objects and staying inside compute and memory limits. We introduce VGTok, a training-free tokenizer that sets patch granularity per region from pixels, ahead of the encoder. VGTok scores each region by morphological top-hat separability from its surround, then thresholds those scores at a per-image percentile, which fixes the token budget. A structure-tensor gate ($\lambda_{\min}$) refines only where two-dimensional object structure supports it, leaving one-dimensional clutter coarse. In a Co-DETR detector with an EVA-02 ViT-L encoder, VGTok exceeds every published VisDrone-val AP and AP$_S$ at every budget from 40\% to 100\% of tokens. At 40\% it records 44.64 AP; the dense detector reaches 48.38 AP, $6.08$ above the strongest published entry. In the same detector with no retraining, VGTok at 70\% of tokens scores $5.7$ AP and $8.0$ AP$_S$ above token merging, and $1.1$ AP$_S$ above a lower input resolution at equal end-to-end GFLOPs. VGTok transfers to AI-TOD-v2 untouched, where the dense detector sets a new state of the art at 37.27 AP and 19.51 AP$_{vt}$. Inserted into a frozen checkpoint, it reaches 36.55 AP at 78.5\% of tokens, above every published entry, where our 376.3M-parameter detector outperforms a 3.0B multi-expert model. We show that a token budget fixed before the backbone, from local separability and structure geometry alone, maintains accuracy on the tiny-object regimes that dominate aerial detection, at $3.1\times$ less encoder compute and $1.9\times$ less encoder memory.
\end{abstract}


\section{Introduction}
\label{sec:intro}

\begin{figure}[t]
  \centering
  \includegraphics[width=\columnwidth]{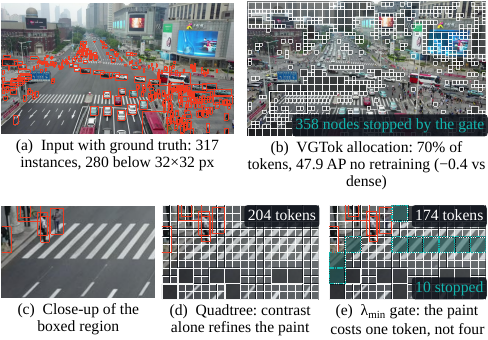}
    \caption{VGTok on a VisDrone frame at $70\%$ tokens.
  Contrast alone spends $204$ tokens on painted crosswalk bars holding no object; the
  $\lambda_{\min}$ gate emits the same crop in $174$.
  (a) Ground truth: $317$ instances, $280$ of them under $32\times32$ px.
  (b) The finest patch size stays on the vehicles while empty roadway, plaza and building
  faces coarsen, at $47.94$ AP with no retraining and no weight changed, $5.64$ AP above the
  strongest published val result; \cref{tab:gate} reports this budget.
  (c) The boxed region, enlarged: paint on plain asphalt.
  (d) Contrast refines that paint to the finest size.
  (e) The gate stops the descent (the recursive splitting of a region), one coarse token in place of four fine ones.}
  \label{fig:teaser}
\end{figure}

A ViT detector reads an image as a grid of patches of one size, and every later stage computes on that grid. Small objects on aerial imagery occupy a handful of pixels, and a grid fine enough to resolve them is too expensive over a full frame. SAHI~\cite{akyon2022sahi} regains that resolution by tiling the frame into overlapping chips, but spends the budget uniformly, giving empty background and dense clusters identical per-pixel allocation. Compute must therefore be spent selectively, based on evidence cheap enough to evaluate before the backbone runs.

Tokenization fixes the token count $N$ before any learned stage, so every mechanism that allocates compute after it (token pruning and merging~\cite{liang2022evit,rao2021dynamicvit,bolya2023tome,fayyaz2022ats,yin2022avit,meng2022adavit}, sparse and density-guided detection heads~\cite{roh2022sparsedetr,zheng2023focusdetr,huang2024dqdetr,wen2026d3rdetr,hu2025domedetr}, crop-and-tile inference~\cite{yang2019clusdet,li2020dmnet,huang2022ufpmpdet,liu2024yolc,zhang2021adazoom}, saliency-guided warping~\cite{recasens2018zoom,thavamani2021fovea,thavamani2023lzu,zheng2025instancewarp}) receives a token budget it cannot enlarge, or leaves $N$ untouched and passes every prediction back through an inverse map.

Any tokenizer that varies the ViT's patch size must decide from pixels alone whether a region deserves resolution, either by learning the decision~\cite{yin2025dart,song2021dge,zhang2023mgvit,chen2023cfvit,havtorn2023msvit} or by thresholding a scalar measure of intra-patch structure (entropy, blur energy, a Laplacian response) at a constant~\cite{ronen2023mixedres,mahajan2026spiralfovea,zhang2024adaptivepatching,qlip2026quadtree}. In both cases the measure scores a region in isolation. Detection needs a different measure. An object is defined by separation from its surround, so the informative quantity is separability from the surround~\cite{shekhar2003unified,gao2007discriminant}. Intra-patch structure and separability diverge where the surround is textured and the target smooth. A scalar average attenuates a target in proportion to its area, so the evidence weakens as the target shrinks. VGTok therefore scores a region, ahead of patch projection, by its separability from the surround, a morphological top-hat whose max over the two polarities keeps the score independent of target area.

Both approaches then refine the same way, replacing a selected region with four sub-patches and emitting all four~\cite{havtorn2023msvit,ronen2023mixedres,yin2025dart}, so refining costs four tokens even where the evidence is a single straight edge. Leaving the regular grid is worse still, giving up the pretrained patch embedding~\cite{aasan2024spit,ziwen2023aff}, and a second uniform scale raises the token count~\cite{chen2022regionvit}. VGTok instead separates the decision to refine from the evidence that would justify it. A region is refined only where $\lambda_{\min}$, the smaller eigenvalue of the structure tensor~\cite{shi1994good}, reports two-dimensional structure, and otherwise stops at its own coarse size (\cref{fig:teaser}). The quadtree has three levels, with patches of $64$, $32$ and $16$ pixels. The token set stays a strict partition. A road marking costs $1$ token where all-or-nothing refinement costs $4$.

We cut tokens with these two training-free statistics in a plain ViT-L detector at $2048\times1152$~\cite{li2022vitdet,zong2023codetr,fang2024eva02}. Every budget from $40\%$ to $100\%$ of the tokens beats every published VisDrone-val AP and AP$_S$: discarding $60\%$ of the sequence leaves $44.64$ AP and $36.03$ AP$_S$, and the same detector run dense reaches $48.38$ AP against the $42.3$ of the strongest published entry~\cite{dronefine2026}, a margin of $6.08$ AP. On AI-TOD-v2, applied with no retuning and no weight changed, VGTok reaches $36.55$ AP at $78.5\%$ of tokens, and $36.42$ at its $70\%$ operating point, against $35.7$ for the strongest entry published on that test split~\cite{scalebridge2025det}, a $3.0$B multi-expert model to our $376.3$M. Inside the same detector, at equal end-to-end GFLOPs and with no retraining, VGTok at $80\%$ of tokens scores at least $0.45$ AP above a lower input resolution on val and $0.30$ on the held-out test-dev split, and $2.84$ AP above ToMe token merging~\cite{bolya2023tome} (\cref{tab:equal}). The shorter sequence runs packed and variable-length~\cite{dehghani2023navit,dong2025flexattention,dao2024flashattention2}, which is what turns removed tokens into removed work~\cite{abdellatif2026dispatch,eliopoulos2025pruning}. Because neither statistic is learned, the budget is not a property of the weights. With a learned scorer, the fraction of tokens kept is fixed once training ends~\cite{havtorn2023msvit,yin2025dart}. With a fixed threshold per scale, that fraction varies from image to image and cannot be chosen in advance~\cite{ronen2023mixedres,zhang2024adaptivepatching,qlip2026quadtree}. VGTok thresholds at a per-image percentile, so the operator sets the fraction directly at inference time. One released checkpoint therefore supports every token budget, so a deployment picks its own point on the accuracy--compute curve and moves it later without retraining.

\section{Related Work}
\label{sec:related}

\subsection{Mixed-scale patchification}
\label{sec:rel:allocation}
\label{sec:rel:granularity}

Aerial tiny-object detection is resolution-limited, since a handful of pixels decides assignment, localisation and evaluation alike~\cite{zhu2022visdrone,wang2021aitod,xu2022nwdrka,xia2018dota,wang2021nwd,xu2022rfla,shi2024simd}. Cropping and image-pyramid methods recover resolution on small objects by running the detector a second time on enlarged regions~\cite{singh2018snip,singh2018sniper,najibi2019autofocus,akyon2022sahi,yang2019clusdet,li2020dmnet,huang2022ufpmpdet,zhang2021adazoom}. Warping methods instead magnify salient regions inside a single resampled image~\cite{recasens2018zoom,thavamani2021fovea,thavamani2023lzu,zheng2025instancewarp}. Sparse and density-guided heads~\cite{carion2020detr,zhu2021deformabledetr,roh2022sparsedetr,zheng2023focusdetr,huang2024dqdetr,wen2026d3rdetr} and token pruning~\cite{liang2022evit,rao2021dynamicvit,hu2025domedetr} then work within a sequence length fixed before they ran.

Patch size is a controllable model axis, and FlexiViT supports many sizes from one weight set~\cite{beyer2023flexivit,dosovitskiy2021vit,fan2024vitar}, uniform across the frame and fixed at deploy time. Varying it by region means recursing on a quadtree, classical in image coding~\cite{horowitz1976picture,samet1984quadtree,sullivan1994quadtree}, where the recursion was never the hard part: exhaustive rate-distortion search was displaced by cheap partition predicates~\cite{bakkouri2020fastcu,hamout2020tensor}. A tokenizer shares neither that search nor its objective, only the demand that the predicate run from pixels before the encoder. (Quadtree attention recurses inside attention, not over granularity~\cite{tang2022quadtree}.)

Several methods already choose the patch size before the backbone runs. Two are training-free: one builds a quadtree and stops subdividing a node when the summed edge magnitude inside it is small, and QLIP inserts a gradient-thresholded quadtree into a frozen encoder~\cite{zhang2024adaptivepatching,qlip2026quadtree}. Others make the same decision from edge, entropy or frequency statistics~\cite{yu2025grcvit,ji2026sepatch3d,aasan2025dht,chen2023cfvit,xie2024quadmamba}. Each stops splitting based on gradient strength, which a painted line produces as readily as a vehicle; we test instead whether the gradients agree on a direction. Quadformer sets an exact token count~\cite{ronen2023mixedres}, MSViT provides per-input adaptivity, learned inside the network~\cite{havtorn2023msvit}, and DART uses a quantile, differentiable inside a learned scorer~\cite{yin2025dart}. A top-down $16/32/64$ patch-size hierarchy over ragged sequences packed block-diagonally is the standard execution path for this family~\cite{havtorn2023msvit,ronen2023mixedres,choudhury2026apt,yin2025dart}. We fix that hierarchy across every tokenizer we compare, so that the scorer is the only term that varies, and re-examine the choice of scorer under a matched budget on aerial data.

Two further choices follow from the aerial setting, and neither is the recursion. An absolute per-scale constant fixes the retained fraction only implicitly, and the constants are themselves tuned~\cite{ronen2023mixedres,zhang2024adaptivepatching,qlip2026quadtree}; a per-image percentile makes that fraction an operator input, gives competing scorers identical budgets, and transfers a rank set on VisDrone to AI-TOD-v2 unchanged. And where adapting a detector to a new tokenizer ordinarily costs a finetune~\cite{havtorn2023msvit,yin2025dart,aasan2025dht}, VGTok's scorer and gate add no parameter and its fusion is zero at initialisation, so the tokenizer can be inserted into a released checkpoint unchanged, which is what the $80\%$ row of \cref{tab:visdrone} is. Refinement stays all or nothing at the node for every tokenizer above, so a region judged worth resolving costs four tokens whether the evidence inside it is a vehicle or a painted line; VGTok tests a busy node (one scoring above the threshold) a second time, on $\lambda_{\min}$, and halts where the gradients agree on a direction. A second family keeps a coarse and a fine token over the same pixels~\cite{chen2022regionvit,wang2021pnp,arta2026,bats2026}. VGTok does not: each pixel belongs to exactly one token.

\subsection{Scoring a region, and running what is left}
\label{sec:rel:signal}
\label{sec:rel:execution}

Each of these tokenizers scores a region by the structure inside it. Quadformer uses pixel blur and two auxiliary forward passes~\cite{ronen2023mixedres}. Space-variant tokenizers use patch entropy as a measure of compressibility~\cite{mahajan2026spiralfovea,schmidt2025stt,jonnalagadda2023foveater}. On aerial frames compressibility and objectness decouple: gravel, foliage and crop texture have high entropy and no objects, while the target is a vehicle on uniform asphalt (\cref{fig:teaser}). Entropy measures intra-patch structure, not separability from the surround, which the centre-surround account of saliency makes central~\cite{kadir2001saliency,bruce2009saliency,gao2007discriminant,itti1998model,achanta2009frequency}.

We take the operator that measures separability from classical morphology. The white and black top-hat is its contrast operator~\cite{serra1982image,sternberg1986grayscale}, multi-scale in Bai and Zhou's infrared form~\cite{bai2010analysis,bai2012image}, whose deep successors reweight features the backbone has already computed~\cite{dai2021alcnet,dai2021acm,xu2026nairstd} while the cheap statistics that precede the encoder rank a uniform grid~\cite{rubab2026dynavit,mahmud2024papr}. The rank is standard too, since an order statistic is the robust threshold of choice~\cite{reed1990adaptive,rohling1983radar,gandhi1988cfar}, and we use it for budget control, which makes the retained fraction an operator input.

A shorter sequence is less work only if it executes as one, and packing variable lengths behind block-diagonal masks is standard practice~\cite{dehghani2023navit,dong2025flexattention,dao2022flashattention,dao2024flashattention2,krell2021packing}; ours follows it. Padding, masking and dispatch can erase a pruning benefit~\cite{bolya2023tome,abdellatif2026dispatch,eliopoulos2025pruning}, so we report FLOPs, wall-clock and peak memory together, and \cref{sec:results:ablation} compares every scorer at one budget under one protocol~\cite{haurum2023which}.

\section{Method}
\label{sec:method}

\begin{figure*}[!t]
  \centering
  \includegraphics[width=0.8\textwidth]{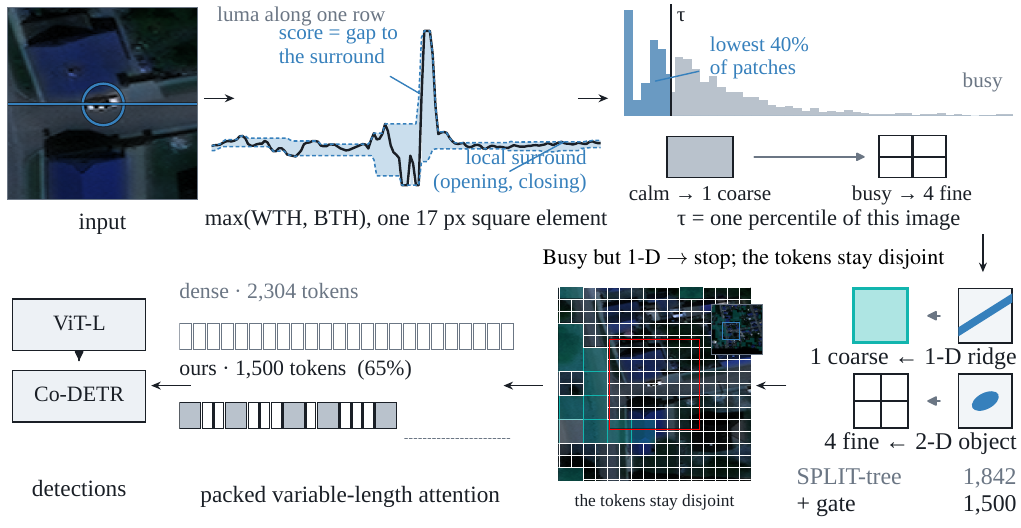}
    \caption{VGTok in four stages: score, split, gate, pack.
  On this frame, at $p=40$ and $r=0.20$, $77$ nodes are gated: the ungated
  descent emits $1842$ tokens, VGTok $1500$ at the same $\tau$.
  Stage $1$ scores every pixel by a top-hat with a $17$-pixel structuring element;
  stage $2$ sets $\tau$ from one per-image percentile and splits the busy nodes;
  stage $3$ stops that descent where $1/(1+\lambda_{\min})$ ranks high, emitting the node at
  its own size and dropping its $4$ children; stage $4$ packs the retained tokens.
  No pixel is described twice.}
  \label{fig:method}
\end{figure*}

We propose VGTok, a tokenizer that replaces the fixed patch grid of a ViT detector with one whose granularity is set per region, from pixels, before any learned stage runs. Both statistics driving VGTok are training-free. The scorer and the gate add no parameter and no auxiliary loss, and they run ahead of the patch projection to set the sequence the backbone receives. A region $v$ receives a score $s(v)$, a threshold $\tau$ separates the regions worth resolving, a rank $r$ decides where refinement stops, and the retained tokens number $N$. Every region is then assigned one of the $3$ patch sizes (\cref{fig:method}).

\subsection{Where to look: separability from the surround}
\label{sec:method:score}

VGTok scores a region by its separability from the surround, using a morphological top-hat~\cite{bai2010analysis,bai2012image}. On BT.601 luma $y$ the white top-hat subtracts from $y$ its opening under a square element of width $w$, leaving what is narrower than $w$, while the black top-hat is the complementary closing. We take both polarities at $w = 17$ and reduce the two responses by a pixelwise maximum. Both polarities are needed because a target can be brighter or darker than its surround: a white car on asphalt responds only to the white top-hat, and a dark car on concrete only to the black one. We set $w=17$ pixels, one more than the finest patch, so that every object smaller than that patch returns its full contrast. Smaller elements add nothing: a top-hat with a larger square is never below one with a smaller square, so the maximum over $w \in \{5,9,17\}$ equals the $w=17$ response. We call this the top-hat score, and write $\mathcal{T}$ for the tokenizer that reads it. That maximum makes $s$ independent of target area for any target narrower than $w$. Objects of side $2$ and side $16$ produce an identical response, a $64$-fold change in area at the same score. Above $w=17$ the opening preserves the object and the response falls to zero. Regions are square patches of $16$, $32$ or $64$ pixels, and a patch takes the strongest response it contains without averaging it away. All $3$ grids are max-pooled from one pixel map on a single canvas padded to a multiple of $64$ pixels, so the grids nest and each coarse patch spans exactly $4$ patches of the next finer grid. Writing $v$ for a node and $v_i$ for its $4$ children, the shared canvas gives $s(v) = \max_i s(v_i)$ exactly.

VGTok sets $\tau$ to the $p$-th percentile of the $16$-pixel grid, per image. The threshold adapts to the image while the retained fraction is set by the operator, so two scorers can be compared at one budget. A matched budget isolates the scorer, since two tokenizers holding the same count spend it on different pixels. A node is calm when $s(v) \le \tau$ and busy otherwise. By the identity, $v$ is busy exactly when at least one of its $4$ children is. We compare two ways of turning the threshold into a token layout: Single-pass ($\mathcal{S}$), used as a reference, and Top-hat ($\mathcal{T}$), the quadtree descent that VGTok uses. Single-pass computes a separate $\tau$ per scale, marks a patch coarse where $s(v) < \tau$, then subtracts every coarsened region from the finer masks. Each scale is decided once and the subtraction leaves the masks disjoint. Top-hat instead takes the single $\tau$ from the finest grid and walks the quadtree top down~\cite{samet1984quadtree,horowitz1976picture}. A calm node is emitted as a leaf and a busy one is discarded in favour of its $4$ children; the masks are disjoint by construction.

\subsection{Where to stop: the $\lambda_{\min}$ gate}
\label{sec:method:gate}

Every quadtree tokenizer in this family spends $4$ tokens refining a node even when one child holds everything worth resolving (\cref{sec:related}). A busy node needs only one busy child, and $3$ of its $4$ are commonly calm. The score cannot indicate which case it is in, because the score is what marked the node busy. On an aerial frame road markings, roof ridges and field boundaries all deviate strongly from their surround. They are contrast and they are not objects. Raising $\tau$ removes them only together with the objects of equal contrast, which on these datasets is most of them. A second threshold on the same score cannot separate them. The gate therefore reads a second statistic, one blind to contrast and so independent of the axis contrast alone cannot decide. What the gate saves grows as the budget tightens. The $4$ tokens a painted line costs are affordable where tokens are plentiful and costly where they are scarce.

What separates an object from a line is the number of directions along which the intensity varies. Let $J$ be the structure tensor of the luma, the outer product of the gradient averaged over a window~\cite{bigun1987optimal,forstner1987fast}, and let $\lambda_{\min}$ be its smaller eigenvalue. Where the gradients inside the window agree on one direction, as they do along an edge or a painted line, $J$ has rank $1$ and $\lambda_{\min} \approx 0$. Where they differ, as they do at a corner or a compact blob, $\lambda_{\min}$ is large. We take the statistic and its interpretation from the classical corner tests: $\lambda_{\min}$ is the test for structure in two directions~\cite{shi1994good}, and the same two eigenvalues form the Harris response~\cite{harris1988combined}. A vehicle is the second case and a lane marking is the first, at equal contrast.

The gate reads a gradient anisotropy score, $g(v) = 1/(1+\lambda_{\min})$, and a node takes the largest $\lambda_{\min}$ it contains. A node is anisotropic, and scores high on $g$, only when it holds no two-dimensional structure anywhere inside it. One vehicle in a node of $64$ pixels is enough to keep $g$ low. We rank the busy nodes Top-hat would visit by $g$, per image, and the top $r$ of them stop splitting, each emitted at its own coarse size with its $4$ children dropped. Every other busy node splits as Top-hat would. The rank is the gate's only setting. We hold it at $r=0.20$ on both datasets. It is an interior optimum bracketed on both sides, and the neighbourhood around it is flat to within $0.07$ AP, which is what lets one value transfer from VisDrone to AI-TOD-v2 untouched. We write the gated tokenizer $\mathcal{T} + g$ where it stands beside its controls, and VGTok for the system as a whole.

New here are the setting and the aggregation: block-level $\lambda_{\min}$ on aerial RGB, under a token budget, scored by detection AP and not encoder speed, ahead of patch projection, where no ViT tokenizer has used the test; the branch it stops is a run of tokens the backbone would otherwise attend over. The link to prior work is deliberate. Deciding how finely to subdivide a block from its structure tensor is established in image and video coding~\cite{bakkouri2020fastcu,hamout2020tensor}, hierarchical edge representations have long held a straight edge in one large leaf~\cite{shneier1981edge,samet1985polygons}, and 3D-HEVC depth coding thresholds this same smaller eigenvalue, crediting the same source we do~\cite{shi1994good}, splitting until only a single directional edge remains~\cite{fu2021cornerpoints}. $\lambda_{\min}$ has a second use, which we build and measure as a control, blended into the pixel score instead of used as a stop rule on the descent. We write that score $\mathcal{T} \oplus \lambda$; it leaves the quadtree unchanged and only reorders which nodes it calls busy. Using $\lambda_{\min}$ in both places gives no further gain, so VGTok applies it once, as the stop rule.

A gated node replaces its children, so the emitted tokens tile the frame exactly once. The measured overlap between tokens is zero at every $r$, and no module is needed to reconcile two descriptions of the same pixels. Written back to the dense grid (\cref{sec:method:exec}), each coarse token fills the $16$-pixel cells it spans, none of them twice. A gated node at $32$ pixels replaces the $4$ base tokens below it, and one at $64$ pixels up to $16$. On the frame of \cref{fig:method}, at $p=40$ and $r=0.20$, $77$ nodes are gated: Top-hat emits $1842$ tokens and VGTok emits $1500$ at the same $\tau$, $18.6\%$ fewer. For Single-pass the retained fraction is a linear function of the percentile. Coarsening the lowest-scoring $p\%$ of patches replaces every four fine tokens with one, so the retained fraction is $1-0.0075\,p$, to within $0.29$ percentage points over the $548$ validation images, and to within $0.03$ from $70\%$ retention upward. This relation holds for Single-pass alone, since a quadtree's token count depends on where the busy nodes sit and the gate moves where they stop. We therefore set quadtree budgets by inverting a measured retention table: $p=81.88$ for Top-hat and $p=74.82$ for the gated tokenizer both reach $40\%$ of the tokens over the same $548$ images. Every comparison we report is calibrated this way.

\subsection{Packed execution}
\label{sec:method:exec}

Tokens at every size are embedded by one convolution applied at the base patch size, so a $64$-pixel token lies in the same space of dimension $D$ as a $16$-pixel one, which a projection per scale would give up. A coarse token is that projection of the resampled patch plus a zero-initialised fusion of the base-size patches it covers~\cite{zhang2023controlnet}. The fusion holds $11.3$M parameters and is exactly zero at initialisation, so our runs without finetuning use the dense weights unchanged and emit predictions identical to the fusion-free path; only the finetuned models train it. Scattering the retained tokens back onto the dense grid leaves every block computing on the full grid. VGTok instead runs all blocks on the packed $(N, D)$ sequence and scatters once, after the last block. The backbone's computation falls as well as the token count~\cite{dehghani2023navit,dong2025flexattention}. Attention is confined to groups of tokens by a single block-diagonal variable-length call~\cite{dao2024flashattention2}, window blocks grouping by window and global blocks by image. It computes the same attention a masked scaled dot-product call would over the same partition, differing only in never materialising the padding, so the pairs a mask would discard are never computed at all. The call uses the xFormers memory-efficient attention kernel; without it the same grouping runs as padded, masked attention (supplementary).

Our detector has three pretraining stages, it is Co-DETR with an EVA-02 ViT-L encoder pretrained by masked image modelling, then given labelled detection pretraining on Objects365 before COCO~\cite{fang2024eva02,zong2023codetr,shao2019objects365}. VGTok uses all three, which is why every tokenizer we compare is run inside this same detector, so the tokenizer is the only term that moves.

\section{Experiments}
\label{sec:results}

We evaluate VGTok on two aerial benchmarks, one dense and one of very small objects. On VisDrone every budget from $40\%$ to $100\%$ of tokens exceeds every published val AP and every published val AP$_S$: at $40\%$ it scores $44.64$ AP and $36.03$ AP$_S$, $+2.34$ and $+3.13$ over published results of $42.3$~\cite{dronefine2026} and $32.9$~\cite{liu2024yolc}, the latter with multi-scale testing, with three fifths of the sequence discarded before the first transformer block. The same tokenizer and rank, applied to AI-TOD-v2 without retuning, set a new state of the art on all three metrics.

\subsection{Setup}
\label{sec:results:setup}

We report VisDrone2019-DET val ($548$ images, $10$ classes) and, in \cref{tab:equal}, its held-out test-dev split ($1{,}610$ images)~\cite{du2019visdronedet}, and AI-TOD-v2 test (the official $14{,}018$-image split, $8$ classes)~\cite{wang2021aitod,xu2022nwdrka}: COCO AP at IoU $0.5$:$0.95$, single model, single scale, no test-time augmentation, VisDrone at maxDets $500$ with ignore regions at IoF $\ge 0.5$ and native $2048\times1152$ input, AI-TOD-v2 at maxDets $1500$ and $2048$, since COCO's default of $100$ understates AP on both. The detector is Co-DETR with an EVA-02 ViT-L encoder~\cite{zong2023codetr,fang2024eva02} pretrained by masked image modelling and then on Objects365 and COCO~\cite{shao2019objects365}, used for every table cell; trained cells are finetuned from the dense checkpoint for $8$ epochs at one image per iteration, so the tokenizer is the only variable.

Budgets are matched by measurement. Each percentile is set by inverting that tokenizer's retention table over all $548$ val images, and the realised token count is re-measured on the same split: every tokenizer is within $0.02$ percentage points of its target, and every comparison is against a control holding at least as many tokens; the supplementary reports the audit. The budget is a mean over the split: at $70\%$ a single image keeps between $61.9\%$ and $77.5\%$ of its tokens. Every trained VisDrone cell in this paper is the mean of three independent finetunes differing only in seed, $6$ in all, at the $40\%$ budget. Differences are compared with a seed floor of $0.15$ AP and $0.14$ AP$_S$, twice the standard error of a difference between two three-seed means pooled over the $2$ replicated cells, and we report the differences that exceed it. Both quadtree tokenizers have the same inherited coarse-patch fusion, so \cref{tab:gate}'s gate effect is a difference between two cells of identical parameter count; Single-pass has one fusion stage fewer and is reported as a reference, never differenced against them. Every row without retraining is one checkpoint, the dense model with the tokenizer swapped in; the trained rows, at $40\%$ on VisDrone and $70\%$ and $40\%$ on AI-TOD-v2, add a finetune at the target budget. Compute is forward-only at batch size $1$ in \texttt{fp16} on a single exclusively held H100 80GB, tabulated in the supplementary.

\begin{figure}[t]
  \centering
  \includegraphics[width=\columnwidth]{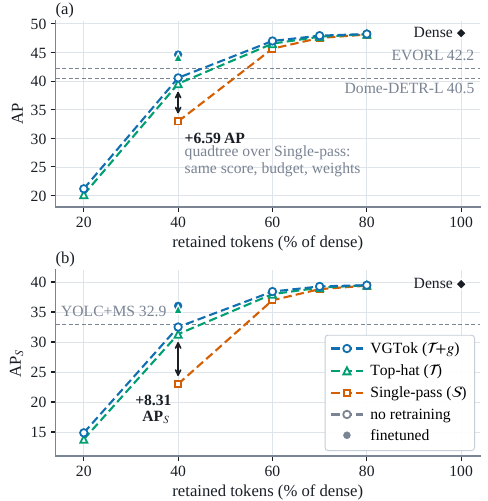}
    \caption{Operating range on VisDrone2019-DET val.
  VGTok exceeds every published result from $40\%$ of tokens upward, finetuned at $40\%$ and with no retraining from $60\%$.
  AP (a) and AP$_S$ (b) against the measured retained-token fraction, $20\%$ to $100\%$ of
  the dense grid; each point is one budget from that tokenizer's measured retention table.
  Dashed, no retraining on the dense checkpoint: VGTok ($\mathcal{T}+g$) over its ungated control Top-hat ($\mathcal{T}$), whose gap is the gate and widens as the budget tightens; Top-hat is worth $+6.59$ AP and $+8.31$ AP$_S$ over Single-pass ($\mathcal{S}$) at $40\%$, from the recursive descent alone (\cref{tab:gate}). Filled markers: VGTok and Top-hat finetuned for $8$ epochs at $40\%$.
  Dotted: EVORL at $42.2$ AP~\cite{zhang2024evoRL}, Dome-DETR-L at
  $40.5$~\cite{hu2025domedetr}, YOLC with multi-scale testing at $32.9$
  AP$_S$~\cite{liu2024yolc}.
  The finetuned points are three-seed means.}
  \label{fig:pareto}
\end{figure}

\subsection{VisDrone2019-DET val}
\label{sec:results:visdrone}
\Cref{tab:visdrone} compares VGTok with the published record on VisDrone2019-DET val, where most prior entries rely on crop-and-zoom pipelines~\cite{huang2022ufpmpdet,zhang2021adazoom}. The highest accuracy here requires native resolution, which VGTok renders practical.
\begin{table}
    \caption{Comparison with the published record on VisDrone2019-DET val. Every token budget from $40\%$ to $100\%$ exceeds every published val AP and every published val AP$_S$, including entries that use multi-scale testing. We reach these numbers at native $2048{\times}1152$ in one forward pass from a single checkpoint, setting the token count at inference; no published entry above sets its own token budget, and most re-run the detector over crops, zooms, or scales. No retraining$^{\ast}$: VGTok inserted into the dense checkpoint at inference, nothing retrained, no head added, no loss changed.}
  \label{tab:visdrone}
  \label{tab:operating}
  \centering
  \scriptsize
  \setlength{\tabcolsep}{2.9pt}
  \begin{tabular}{@{}lllrrr@{}}
    \toprule
    Method & Backbone & Input & AP & AP$_{50}$ & AP$_S$ \\
    \midrule
    DroneFINE-L~\cite{dronefine2026} & Swin-L (G-DINO) & concurrent & 42.3\phantom{0} & 65.8\phantom{0} & --- \\
    EVORL~\cite{zhang2024evoRL} & CSPDarkNet & multi-crop & 42.2\phantom{0} & 66.0\phantom{0} & --- \\
    Focus-and-Detect~\cite{focusdetect2022} & ResNeXt-101 & 2-stage crop & 42.06 & 66.12 & 32.0\phantom{0} \\
    Dome-DETR-L~\cite{hu2025domedetr} & HGNetv2-B4 & $1200{\times}800$ & 40.5\phantom{0} & 64.5\phantom{0} & --- \\
    AdaZoom~\cite{zhang2021adazoom} & ResNeXt-101 & multi-crop & 40.3\phantom{0} & 66.9\phantom{0} & --- \\
    UFPMP-Det\,+\,MS~\cite{huang2022ufpmpdet} & ResNeXt-101 & crop\,+\,MS & 40.1\phantom{0} & 66.8\phantom{0} & --- \\
    YOLC\,+\,MS~\cite{liu2024yolc} & HRNet & crop\,+\,MS & 39.6\phantom{0} & 63.7\phantom{0} & 32.9\phantom{0} \\
    Dome-DETR-L~\cite{hu2025domedetr} & HGNetv2-B4 & $800^2$ & 39.0\phantom{0} & 61.1\phantom{0} & --- \\
    YOLC~\cite{liu2024yolc} & HRNet & crop & 37.8\phantom{0} & 61.7\phantom{0} & 30.5\phantom{0} \\
    \midrule
    Dense, $100\%$ tokens & \multicolumn{1}{c}{\multirow{4}{*}{ViT-L (EVA-02)}} & \multicolumn{1}{c}{\multirow{4}{*}{$2048{\times}1152$}} & \textbf{48.38} & \textbf{74.28} & \textbf{39.63} \\
    VGTok, $80\%$ tokens$^{\ast}$ & & & 48.23 & 74.16 & 39.47 \\
    VGTok, $70\%$ tokens$^{\ast}$ & & & 47.94 & 73.85 & 39.26 \\
    \emph{VGTok, $40\%$ tokens} & & & 44.64 & 70.01 & 36.03 \\
    \bottomrule
  \end{tabular}
\end{table}

The $80\%$ entry without retraining is VGTok inserted into our dense checkpoint at inference, with no head added, no loss changed and no weight touched. It reaches $48.23$ AP, $5.93$ above the strongest published val result and within $0.15$ of our own dense result ($48.38$), so a fifth of the sequence can be discarded at deployment time. A fifth is also loose enough that the allocator is not the limiting factor, since a flat per-scale percentile differs by less than the seed floor here ($48.13$ against our measured $48.23$), which is why the allocator is measured where budgets are tight (\cref{sec:results:ablation}).

We adopt $40\%$ tokens as the operating point. It is the lowest budget that still exceeds every published AP and AP$_S$, and the regime where the tokenizer's selection separates from its control by more than three times the seed floor. At this budget three-fifths of the dense grid is discarded, yet the model records $44.64$ AP and $36.03$ AP$_S$, $2.34$ AP and $3.13$ AP$_S$ above the strongest published entries (\cref{tab:visdrone}). The two metrics fall nearly in step ($3.50$ and $3.50$ below a dense model finetuned under the identical $8$-epoch recipe, $48.14$ AP and $39.53$ AP$_S$), and that near-parity is the signature of content-aware pruning. A uniform cut of this depth would remove small-object tokens first and open a larger gap on AP$_S$. VGTok instead coarsens the empty nodes and assigns the remaining tokens to the finest scale (\cref{fig:teaser}), so a $2.5\times$ reduction leaves the object-bearing tokens largely intact. At the looser $70\%$ and $80\%$ budgets the dense checkpoint already retains the accuracy.  The $40\%$ column is where the allocation matters (\cref{sec:results:ablation}).

\begin{table}[t]
  \caption{One detector and one checkpoint on VisDrone2019-DET, no retraining: the
  dense detector at lower input resolutions, VGTok, and ToMe~\cite{bolya2023tome}.
  GFLOPs and latency are end to end, as in \cref{tab:compute}. Test-dev is held out.}
  \label{tab:equal}
  \centering
  \scriptsize
  \setlength{\tabcolsep}{2.1pt}
  \begin{tabular}{@{}lrrrrrrrr@{}}
    \toprule
    & & & & \multicolumn{3}{c}{val} & \multicolumn{2}{c}{test-dev} \\
    \cmidrule(lr){5-7}\cmidrule(l){8-9}
    & Tok.\,\% & GFLOPs & ms & AP & AP$_{50}$ & AP$_S$ & AP & AP$_S$ \\
    \midrule
    \multicolumn{9}{@{}l}{\emph{Dense, input long side}} \\
    \quad $2048$ px & $100$ & $12{,}956$ & $179.8$ & $48.38$ & $74.28$ & $39.63$ & $37.61$ & $26.38$ \\
    \quad $1920$ px & $100$ & $11{,}160$ & $160.8$ & $47.86$ & $73.61$ & $39.02$ & $37.08$ & $25.70$ \\
    \quad $1792$ px & $100$ & $9{,}838$ & $145.8$ & $47.33$ & $72.92$ & $38.13$ & $36.61$ & $25.02$ \\
    \quad $1664$ px & $100$ & $8{,}645$ & $133.5$ & $46.50$ & $71.98$ & $37.23$ & $36.14$ & $24.18$ \\
    \quad $1536$ px & $100$ & $7{,}005$ & $113.6$ & $45.52$ & $71.09$ & $35.97$ & $35.45$ & $23.54$ \\
    \multicolumn{9}{@{}l}{\emph{VGTok, $2048$ px}} \\
    \quad $80\%$ & $80.0$ & $10{,}740$ & $155.8$ & $48.23$ & $74.16$ & $39.47$ & $37.39$ & $26.21$ \\
    \quad $70\%$ & $70.0$ & $9{,}822$ & $152.5$ & $47.94$ & $73.85$ & $39.26$ & $36.91$ & $25.83$ \\
    \quad $60\%$ & $60.0$ & $8{,}953$ & $146.6$ & $47.01$ & $72.91$ & $38.41$ & $36.09$ & $25.28$ \\
    \quad $40\%$ & $40.0$ & $7{,}345$ & $136.3$ & $40.56$ & $65.69$ & $32.50$ & $30.94$ & $21.69$ \\
    \multicolumn{9}{@{}l}{\emph{ToMe, $2048$ px}} \\
    \quad $m=10$ & $79.3$ & $10{,}749$ & $183.0$ & $45.38$ & $71.26$ & $35.50$ & $35.46$ & $23.42$ \\
    \quad $m=14$ & $70.7$ & $9{,}948$ & $175.7$ & $42.25$ & $67.72$ & $31.30$ & $33.10$ & $20.44$ \\
    \bottomrule
  \end{tabular}
\end{table}

The published entries of \cref{tab:visdrone} differ from ours in backbone and input size, so \cref{tab:equal} compares inside one detector and one checkpoint, with no retraining: the dense detector at a lower input resolution, and ToMe token merging~\cite{bolya2023tome}. At equal end-to-end GFLOPs, interpolating between the dense input sizes, VGTok at $80\%$ of tokens scores $0.54$ AP and $0.73$ AP$_S$ above the lower resolution on val, and $0.36$ and $0.57$ on the held-out test-dev split. At $70\%$ it scores $0.62$ AP and $1.14$ AP$_S$ above it on val and $0.57$ AP$_S$ on test-dev, where the AP difference is inside its interval. It scores $2.84$ AP above ToMe at $80\%$, and $5.69$ AP and $7.96$ AP$_S$ above it at $70\%$, at equal or lower GFLOPs. Below $60\%$ the comparison reverses without fine-tuning: at $40\%$ VGTok scores $40.56$ AP where the dense detector at $1536$ px scores $45.52$ at similar cost. The dense rows use the detector's default attention; on the packed attention that VGTok uses they are $8$ to $11\%$ faster, the val margins at $80\%$ become $0.45$ AP and $0.59$ AP$_S$, and VGTok has no latency advantage at equal accuracy; the supplementary gives these rows and every interval.

\subsection{AI-TOD-v2 test}
\label{sec:results:aitod}

\Cref{tab:aitod} shows that the same recipe transfers.
The tokenizer and rank $r=0.20$ are transferred from VisDrone to AI-TOD-v2 without retuning and evaluated on objects an order of magnitude smaller. At full retention the dense detector reaches 37.27 AP and 19.51 AP$_{\mathrm{vt}}$, a new state of the art on all three, $+1.57$ AP and $+4.38$ AP$_{50}$ over ScaleBridge-Det~\cite{scalebridge2025det} and $+0.51$ AP$_{\mathrm{vt}}$ over Dome-DETR-L~\cite{hu2025domedetr}. The tokenizers at $70\%$ and above hold that lead on AP and AP$_{50}$. VGTok at 78.5\% of tokens, the loosest budget the gate reaches at $r = 0.20$, exceeds every previously published AP ($36.55$ against $35.7$) with no weight changed, outperforming a $3.0$B multi-expert model by $0.85$ AP with a $376.3$M detector, an eighth of the parameters. Detecting objects this small requires resolution. We process these $800^2$ frames at $2048^2$, against $800^2$ to $1024^2$ for the published entries. Making that resolution affordable is what VGTok is for, and the supplementary shows the accuracy cost of a fixed token cut only falls as the input grows.

\begin{table}
\caption{Transfer to AI-TOD-v2 test, no retuning. At $100\%$ of tokens the dense detector is best in every column: $37.27$ AP against ScaleBridge-Det's $35.7$, $76.48$ AP$_{50}$ against its $72.1$, and $19.51$ AP$_{\mathrm{vt}}$ against Dome-DETR-L's $19.0$. The tokenizer and rank $r=0.20$ are transferred from VisDrone under the same four budgets; protocol as in \cref{sec:results:setup}, and AP$_{\mathrm{vt}}$ is the very-tiny band ($<8^2$\,px). Italics mark our operating point, $70\%$ here against VisDrone's $40\%$. No retraining$^{\ast}$: no weight changed. $78.5\%$ is the loosest budget the gate reaches at $r=0.20$.}
  \label{tab:aitod}
  \centering
  \scriptsize
  \setlength{\tabcolsep}{2.9pt}
  \begin{tabular}{@{}lllrrr@{}}
    \toprule
    Method & Backbone & Input & AP & AP$_{50}$ & AP$_{\mathrm{vt}}$ \\
    \midrule
    ScaleBridge-Det (3.0B)~\cite{scalebridge2025det} & MoE 3.0B     & $1024^2$ & 35.7\phantom{0} & 72.1\phantom{0} & 16.2\phantom{0} \\
    Dome-DETR-L~\cite{hu2025domedetr}                & HGNetv2-B4   & $800^2$  & 34.6\phantom{0} & 69.2\phantom{0} & 19.0\phantom{0} \\
    Dome-DETR-M~\cite{hu2025domedetr}                & HGNetv2-B2   & $800^2$  & 34.0\phantom{0} & 68.4\phantom{0} & 18.4\phantom{0} \\
    ScaleBridge-Det (2.0B)~\cite{scalebridge2025det} & MoE 2.0B     & $1024^2$ & 33.6\phantom{0} & 71.8\phantom{0} & 14.2\phantom{0} \\
    Dome-DETR-S~\cite{hu2025domedetr}                & HGNetv2-B0   & $800^2$  & 33.3\phantom{0} & 67.5\phantom{0} & 17.8\phantom{0} \\
    D$^3$R-DETR~\cite{wen2026d3rdetr}                & HGNetv2-B0   & $800^2$  & 31.3\phantom{0} & 65.1\phantom{0} & 16.6\phantom{0} \\
    DQ-DETR~\cite{huang2024dqdetr}                   & ResNet-50    & $800^2$  & 30.2\phantom{0} & 68.6\phantom{0} & 15.3\phantom{0} \\
    RFLA~\cite{xu2022rfla}                           & DetectoRS    & $800^2$  & 25.7\phantom{0} & 58.9\phantom{0} & 9.3\phantom{0}  \\
    NWD-RKA~\cite{xu2022nwdrka}                      & DetectoRS    & $800^2$  & 24.7\phantom{0} & 57.2\phantom{0} & 9.7\phantom{0}  \\
    \midrule
    Dense, $100\%$ tokens & \multicolumn{1}{c}{\multirow{4}{*}{ViT-L (EVA-02)}} & \multicolumn{1}{c}{\multirow{4}{*}{$2048^2$}} & \textbf{37.27} & \textbf{76.48} & \textbf{19.51} \\
    VGTok, $78.5\%$ tokens$^{\ast}$                  &  &  & 36.55 & 75.69 & \textbf{19.38} \\
    \emph{VGTok, $70\%$ tokens}                      &  &  & 36.42 & 73.76 & 18.59 \\
    VGTok, $40\%$ tokens                             &  &  & 34.62 & 72.01 & 16.45 \\
    \bottomrule
  \end{tabular}
\end{table}

The supplementary reports the gate ablation band by band. Without retraining, the gate is
positive on every band that has the instances to measure it: on aggregate AP it is
worth $+0.29$, $+0.13$ and $+0.31$ at $70\%$, $40\%$ and $20\%$ of tokens, and on the
tiny band, which holds $275{,}050$ of the split's $376{,}121$ instances, it is
positive and flat, $+0.35$, $+0.34$, $+0.38$. Only AP$_{\mathrm{vt}}$ changes sign,
and that band is an unweighted mean over eight classes of which four hold $105$ of
its $48{,}995$ boxes, so half of it rests on a fifth of a percent of the annotations.
The gate effect on this benchmark is assessed from the bands and classes that do resolve it. The per-class results show the mechanism. The gate moves five of the eight classes up, and three of the four classes holding more than a thousand very-tiny boxes move up.

What the dataset does change is how far the budget can be pushed. The gate gains by not descending on one-dimensional structure, so its effect tracks the
composition of the population the descent still reaches. On VisDrone the clutter
fraction among gate-visible nodes falls from $79.1\%$ to $56.1\%$ between the
$70\%$ and $20\%$ budgets and the gate effect grows with it, from $+0.18$ AP without retraining at
$70\%$ to $+1.02$ at $40\%$ and $+1.02$ at $20\%$. On AI-TOD it stays above $91\%$ and the effect stays flat, before training and after
it. We attribute this to what the descent reaches rather than to the statistic, which
separates clutter from objects better on AI-TOD than on VisDrone, as the
node-level audit in the supplementary shows. The
range we report is accordingly $70$ to $100\%$ here and $40$ to $100\%$ on VisDrone,
fixed at the lower end by the trained rows of \cref{tab:aitod}. A criterion that
predicts where its own effect will be flat, and is flat there, is tested by more than
its gains.

\subsection{From tokens to FLOPs}
\label{sec:results:compute}
\begin{table}
    \caption{Forward-pass cost at four token budgets. Packing is what turns a token cut into a
  compute cut: the scatter control holds the same gate and the same retained tokens but returns
  them to the dense grid before the first block, and costs more than dense.
  Batch $1$, \texttt{fp16} with the variable-length kernel, one exclusively held H100 80GB,
  median over three repeats on all $548$ val images with the dense checkpoint loaded; every
  end-to-end budget is re-measured and is within $0.0003$ of its target. The memory column and the control row come from a second run of the same protocol. }
  \label{tab:compute}
  \centering
  \small
  \setlength{\tabcolsep}{3.5pt}
  \begin{tabular}{@{}lrrrrrr@{}}
    \toprule
    & & \multicolumn{3}{c}{Encoder} & \multicolumn{2}{c}{End-to-end} \\
    \cmidrule(lr){3-5}\cmidrule(l){6-7}
    Tokenizer & Tokens & GFLOPs & ms & GB & GFLOPs & ms \\
    \midrule
    Dense & $100.0\%$ & 8281 & 92.4 & 0.466 & 12955.5 & 179.8 \\
    VGTok & $80.0\%$ & 6066 & 68.2 & 0.409 & 10740.3 & 155.8 \\
    VGTok & $70.0\%$ & 5148 & 64.7 & 0.372 & 9821.8 & 152.5 \\
    \emph{VGTok} & $40.0\%$ & 2671 & 48.2 & 0.251 & 7344.8 & 136.3 \\
    \midrule
    Control & $70.0\%$ & 8292 & 104.6 & 0.498 & 12966.3 & 192.7 \\
    \bottomrule
  \end{tabular}
\end{table}
VGTok turns a token reduction into a measured compute reduction.
At $70\%$ of the tokens the encoder costs $5148$ GFLOPs against the dense $8281$; at $40\%$ it
costs $2671$, a $3.1\times$ reduction. Latency falls from
$92.4$ to $48.2$\,ms and peak memory from $0.466$ to $0.251$\,GB, and end to end the detector
drops from $12955.5$ to $7344.8$ GFLOPs and from $179.8$ to $136.3$\,ms (\cref{tab:compute}). The tokenizer itself takes $12.9$ to $14.9$\,ms per image; at $80\%$ of tokens that is $4.2$\,ms for the top-hat score, $2.6$ for $\lambda_{\min}$, $1.3$ for the quadtree and the gate, $3.9$ for packing and $0.9$ for the final scatter (supplementary).
Encoder cost drops faster than the token count because the
packed path evaluates attention only within its groups, where the dense grid evaluates it over the
whole frame.

The reduction is possible only because VGTok emits a strict partition. The retained tokens
form a single variable-length sequence rather than a dense grid with holes. Every block
therefore runs on the packed $(N,D)$ sequence and scatters once after the last~\cite{dehghani2023navit,dong2025flexattention}, with attention confined to its groups by one
block-diagonal variable-length call~\cite{dao2024flashattention2}. Padding is neither stored
nor computed.

A scatter control isolates the contribution of that execution path. It keeps the identical
gate and the identical retained tokens, then returns them to the dense grid before the first
block. The control spends $8292$ GFLOPs against dense’s $8281$, with higher latency and higher
memory. Discarding $30\%$ of the tokens without packing is strictly worse than discarding
none. Same tokenizer and same tokens produce opposite outcomes: the compute saving requires
the packed path.

The encoder is the whole of what a tokenizer can affect, and VGTok reduces its cost by $3.1\times$. The neck
and Co-DETR head read a dense feature map whose shape is fixed by input resolution and the decoder’s
query set, not by $N$, so their cost is a constant of the detector under every tokenizer and every
budget: encoder peak memory falls by $1.9\times$, and the $4.54$--$4.65$\,GB that remains is used by the head, not by the tokenizer. The scorer and the gate add no
parameters. The rise from $365.0$M to $376.3$M is entirely the inherited patch embedding. Head-side
sparsity and token-side sparsity operate in different domains and compose directly.

\subsection{Ablation study}
\label{sec:results:ablation}

We first ask how much freedom there really is in the choice of pixel score, holding the token allocator and budget fixed. Under a common $70\%$ token budget with Single-pass, we evaluate eight training-free pixel scores. Their ranking without retraining spans $1.94$ AP and $2.34$ AP$_S$, so the signal a tokenizer reads is worth that much on its own, before any allocator decides anything; the supplementary reports the full ordering, and the same comparison inside the gated quadtree, where at $40\%$ of tokens the top-hat is $4.53$ AP above patch entropy and $2.32$ above gradient energy. The two statistics VGTok reads take the first two places of the eight, separated by $0.08$ against a $0.15$ floor, and the top-hat we use sits $1.86$ AP above patch entropy, the intra-patch detail measure this family of tokenizers uses (\cref{sec:related}). We use the top-hat because the same statistic is worth more used at the gate than in the score, which is what the rest of this section measures. Every entry in that ranking shares one limitation. Every one is a contrast measure, and contrast on an aerial frame is as often a road marking or a field boundary as an object. One-dimensional structure is not noise a better threshold removes; it is exactly what a contrast score is defined to find, and at tight budgets it competes directly with objects for tokens.

The $\lambda_{\min}$ gate is designed to make the distinction no contrast measure can. It acts on the quadtree itself rather than on the score. We halt descent wherever the gradients in a node already agree on one direction, and we keep the token set a strict partition of the image (\cref{fig:method}). In other words, the gate treats long, thin structures as regions that should not be further subdivided, preventing the allocator from spending tokens on repeated contrast along the same one-dimensional support. \cref{tab:gate} measures the gate exactly where no contrast scorer can substitute for it. It intervenes when there is a budget trade-off to make, and its effect grows as that trade-off tightens. The gate also loses objects: at $70\%$ of tokens an object centred in a token the gate stopped is missed $13\%$ of the time, against $0.5\%$ in a $16$-pixel patch, and the miss rate rises with aspect ratio up to $4$; $81\%$ of the objects VGTok misses nevertheless lie in $16$-pixel patches (supplementary).

\begin{table}
\caption{Gate ablation, VisDrone2019-DET val, five matched budgets.
The $\lambda_{\min}$ gate's effect grows as the budget tightens, reaching $+1.02$ AP at $40\%$ and $20\%$ and $+1.18$ AP$_S$ at $40\%$; finetuned at the $40\%$ operating point it is $+0.48$ AP and $+0.59$ AP$_S$, and the three finetunes per cell do not overlap, $[44.55,\,44.75]$ against $[44.10,\,44.25]$. The run begins at $+0.04$ AP and $+0.01$ AP$_S$ at $80\%$,
where the budget is loose.
AP above, AP$_S$ below, no retraining, every cell one evaluation of the dense checkpoint
(\cref{sec:results:setup}).
The two $\mathcal{T}$ rows differ only in the gate at rank $r = 0.20$, and \emph{gate effect}
is their difference. \emph{unr.} marks budgets Single-pass cannot reach.
Bold compares the two $\mathcal{T}$ rows, which differ only in the gate, and marks only
differences that exceed the seed floor.}

  \label{tab:gate}
  \centering
  \footnotesize
  \setlength{\tabcolsep}{4pt}
  \newcommand{\gcw}{0.1096\linewidth}%
  \newcommand{\gn}[1]{\makebox[\gcw][r]{#1}}%
  \newcommand{\gs}[1]{\makebox[\dimexpr0.452\linewidth-12\tabcolsep\relax][l]{#1}}%
  \begin{tabular}{@{}llllll@{}}
    \toprule
    \gs{Tokenizer (AP)} & \gn{$80\%$} & \gn{$70\%$} & \gn{$60\%$} & \gn{$40\%$} & \gn{$20\%$} \\
    \midrule
    \gs{Single-pass ($\mathcal{S}$)}    & \gn{48.13} & \gn{47.54} & \gn{45.67} & \gn{32.96} & \gn{\emph{unr.}} \\
    \addlinespace[3pt]
    \gs{Top-hat ($\mathcal{T}$)}        & \gn{48.18} & \gn{47.76} & \gn{46.54} & \gn{39.54} & \gn{20.17} \\
    \gs{Top-hat ($\mathcal{T}$) $+\,g$} & \gn{48.23} & \gn{\textbf{47.94}} & \gn{\textbf{47.01}} & \gn{\textbf{40.56}} & \gn{\textbf{21.18}} \\
    \cmidrule(l){2-6}
    \gs{\emph{gate effect}}             & \gn{$+0.04$} & \gn{$+0.18$} & \gn{$+0.47$} & \gn{$+1.02$} & \gn{$+1.02$} \\
    \midrule
    \gs{Tokenizer (AP$_{S}$)} & \gn{$80\%$} & \gn{$70\%$} & \gn{$60\%$} & \gn{$40\%$} & \gn{$20\%$} \\
    \midrule
    \gs{Single-pass ($\mathcal{S}$)}    & \gn{39.39} & \gn{38.81} & \gn{36.96} & \gn{23.00} & \gn{\emph{unr.}} \\
    \addlinespace[3pt]
    \gs{Top-hat ($\mathcal{T}$)}        & \gn{39.46} & \gn{39.05} & \gn{37.98} & \gn{31.31} & \gn{13.79} \\
    \gs{Top-hat ($\mathcal{T}$) $+\,g$} & \gn{39.47} & \gn{\textbf{39.26}} & \gn{\textbf{38.41}} & \gn{\textbf{32.50}} & \gn{\textbf{14.86}} \\
    \cmidrule(l){2-6}
    \gs{\emph{gate effect}}             & \gn{$+0.01$} & \gn{$+0.21$} & \gn{$+0.43$} & \gn{$+1.18$} & \gn{$+1.07$} \\
    \bottomrule
  \end{tabular}
\end{table}

VGTok adds one statistic and uses it once, at the gate. Nothing of our own is trained or tuned. Blending $\lambda_{\min}$ into the pixel score instead is worth $+0.70$ AP at the $40\%$ budget against $+1.02$ for the gate, and using it both ways is worse than the gate alone at $40\%$ and no better at any other budget. The separation at that budget is not seed noise: the three finetunes behind $44.64$ span $[44.55,\,44.75]$ and the three behind the ungated control span $[44.10,\,44.25]$, two ranges that do not touch on AP, and the same holds on AP$_S$.

Depth and the gate both matter at tight budgets, and we isolate each to show this. The Single-pass tokenizer (no recursion, no gate) is run at every budget it can reach, and reading that row across is the clearest statement of the thesis we can make. Its operating range ends where ours begins to matter. Two levels limit retention to at least a quarter of the dense grid, so no threshold reaches a $20\%$ budget at any setting, and at a matched $40\%$ without retraining the recursive descent alone is worth $+6.59$ AP over it. Across the range they share, Single-pass and VGTok track each other to within half a point down to $70\%$ ($47.54$ against $47.94$), then separate: at $40\%$ Single-pass falls $7.61$ AP behind ($32.96$ against $40.56$). The crossing is at a measured budget, not an interpolation between other budgets. Where tokens are plentiful no allocator is needed and the two score about the same; the gated three-level descent is built for the range where allocation matters, and it is the only tokenizer we test that reaches $20\%$ at all.

Reading the no-gate row across shows why the tight end of this range matters most: $80\%$ to $60\%$ costs $1.64$ AP, $60\%$ to $40\%$ costs $7.00$, and $40\%$ to $20\%$ costs $19.38$ as the budget begins coarsening the objects themselves. The gate is worth its largest margins there, $+1.02$ AP at $40\%$ and $20\%$ and $+1.18$ AP$_S$ at $40\%$.

\section{Conclusion}
\label{sec:conclusion}

The room for improvement in adaptive tokenization lies in the criterion, not the mechanism.
VGTok tests it with two training-free statistics read from pixels ahead of the
encoder. A top-hat scores each region by separability from its surround, and a
$\lambda_{\min}$ gate halts refinement where the gradients agree on a
direction. At every budget from 40 to 100\% of the dense grid it exceeds all
published VisDrone-val AP and AP$_S$, where the dense detector reaches 48.38~AP,
6.08 above the strongest published entry. Inside one detector at equal end-to-end GFLOPs it scores above a lower input resolution at 80\% of tokens, on val and on the held-out test-dev split, and above token merging. It transfers to AI-TOD-v2 untouched,
to a state of the art on all three metrics, where 376.3M parameters outperform a
3.0B multi-expert model.

The criterion accounts for these results. With the allocator, the patch-size hierarchy and the
budget fixed, eight training-free statistics span 1.94~AP, and patch entropy,
the measure this family of tokenizers reads, sits 1.86 below ours. What a
tokenizer looks for is worth more than how it descends, and more the less there
is to spend. The gate is worth 1.02~AP where tokens are scarcest and 0.04
where they are plentiful, and the same explanation predicts where it should be
flat. On AI-TOD-v2, where the clutter fraction stays high, it is flat. Because
no part of the tokenizer is learned, one checkpoint supports every token budget and
the budget stays an operator input after training.

\section*{Acknowledgment}
The author used Claude (Anthropic) to edit the text of all sections and to write plotting and evaluation scripts; the author checked all text, code and results.

\bibliographystyle{IEEEtran}
\bibliography{main}

\clearpage
\onecolumn
\floatsintextorder
\appendices
\section*{Supplementary Material}

\setcounter{section}{0}
\setcounter{table}{0}
\setcounter{figure}{0}
\renewcommand{\thesection}{\Alph{section}}
\renewcommand{\thetable}{A\arabic{table}}
\renewcommand{\thefigure}{A\arabic{figure}}

In this supplementary material, we provide the analyses and protocol details deferred from \cref{sec:results}. \Cref{sec:suppl:auroc} isolates the gate statistic at the level of individual quadtree nodes: $\lambda_{\min}$ ranks clutter above object-bearing structure at an AUROC of $0.712$ to $0.750$, where four contrast-derived alternatives reach $0.52$ to $0.65$. \Cref{sec:suppl:scorers} then ablates eight training-free pixel scores at a matched budget. \Cref{sec:suppl:ladders} separates the splitter from the gate, where the recursive descent alone is worth $+6.59$ AP at a matched $40\%$ budget. \Cref{sec:suppl:bands,sec:suppl:aitod} decompose the gain by object scale on VisDrone and by area band and class on AI-TOD-v2. The gate moves the small-object band further than the aggregate, $+0.59$ AP$_S$ against $+0.48$ AP at the $40\%$ operating point. \Cref{sec:suppl:rank,sec:suppl:dense,sec:suppl:res} report sensitivity to the gate rank $r$, the convergence of the dense reference, and sensitivity to input resolution. \Cref{sec:suppl:seeds,sec:suppl:budget} give the seed protocol behind the seed floor and the token-budget audit behind every matched comparison, and \cref{sec:suppl:failure} bounds how much of what the gate declines to refine was describing an object.

Two conventions govern the whole document, and we state them here because the sections that use them come first. Every difference we argue for clears the seed floor, $0.15$ AP and $0.14$ AP$_S$, pooled over $2$ trained VisDrone cells and $6$ finetunes (\cref{sec:suppl:seeds}). We set every token budget by inverting that tokenizer's own measured retention table and then re-measure the result, so we hold every tokenizer to the same token count as the control it is compared against (\cref{sec:suppl:budget}). The four tokenizers are named as in \cref{sec:method}. $\mathcal{T}$ is driven by the top-hat score. Adding $+\,g$ puts the $\lambda_{\min}$ gate on the descent at rank $r=0.20$, ranking nodes by gradient anisotropy, and $\oplus\,\lambda$ blends $\lambda_{\min}$ into the pixel score instead.

Every cell below is our own measurement, so no table carries a quotation marker or an unrun cell. One symbol appears: \emph{unreachable} is a cell no threshold can reach, because at the shipped rank the gate alone caps retention below the stated budget. That is a structural property of the tokenizer, not a gap in the sweep. Bold marks the best value in a column among comparable rows and carries no other meaning. We never use it to mark the shipped setting, and never to mark a difference, since a difference is not a value that can be best. A column whose top two values sit closer than the seed floor carries no bold at all, because a lead the noise floor cannot resolve is not a lead.

\section{Node-level discrimination of \texorpdfstring{$\lambda_{\min}$}{lambda-min}}
\label{sec:suppl:auroc}

$\lambda_{\min}$ tells one-dimensional structure (road markings, roof ridges, field boundaries) from the two-dimensional structure of an object, and contrast alone cannot. That claim is falsifiable at the level of individual nodes, so we test it there. \Cref{tab:suppl:auroc} puts the shipped statistic against four contrast-derived alternatives on both datasets, then follows it down the budget range.

We fix the population by the ungated quadtree's own descent: every node that is busy, reachable, and not padding, at the two levels where a gate can act ($32$ and $64$ pixels). We label a node \emph{clutter} if no ground-truth box overlaps it. We then ask how well each candidate statistic ranks clutter above object-bearing nodes and report the area under the ROC curve. Fixing the population under the ungated descent matters: a statistic that fired at $64$ pixels would otherwise hide its own $32$-pixel children and change the denominator under itself.

\paragraph{Which images these are.} Both pools are $150$ images. On VisDrone we draw them from the $548$-image val split, the split every VisDrone AP in this paper is scored on, so the audit and the accuracy it explains sit on the same images. On AI-TOD-v2 we draw them from the $2{,}804$-image \emph{val} split. There the audit is an out-of-sample check: its images are disjoint from the $14{,}018$-image \emph{test} split the AP tables score, so it measures what the statistic separates on the benchmark rather than on the evaluation images themselves.

\begin{table}
    \caption{Gate statistic discrimination, with a budget sweep. AUROC for ranking clutter nodes, those no ground-truth box overlaps, above object-bearing ones: only $1/(1+\lambda_{\min})$ clears chance meaningfully, at $0.712$ or better in every column. Two levels and not three: the gate is a stop rule, so it acts at $64$ and $32$\,px, and the $16$\,px leaf has no split to stop, so that column would be structurally empty, not unrun. Pools are $150$ VisDrone \emph{val} and $150$ AI-TOD-v2 \emph{val} images. Bold is the best of the five candidates in an upper-block column, fixed at $70\%$.}
  \label{tab:suppl:auroc}
  \label{tab:suppl:aurocbudget}%
  \centering
  \footnotesize
  \setlength{\tabcolsep}{4pt}
  \begin{tabular}{@{}lrrrr@{}}
    \toprule
    & \multicolumn{2}{c}{VisDrone} & \multicolumn{2}{c}{AI-TOD-v2} \\
    \cmidrule(lr){2-3}\cmidrule(lr){4-5}
    & 32\,px & 64\,px & 32\,px & 64\,px \\
    \midrule
    \multicolumn{5}{@{}l}{\emph{Five candidates, at the $70\%$ budget}} \\
    $1/(1+\lambda_{\min})$    & \textbf{0.712} & \textbf{0.731} & \textbf{0.748} & \textbf{0.750} \\
    node vs.\ surround $\times$ homogeneity & 0.599 & 0.639 & 0.612 & 0.648 \\
    within-node homogeneity   & 0.602 & 0.624 & 0.606 & 0.634 \\
    node vs.\ surround        & 0.552 & 0.601 & 0.580 & 0.616 \\
    fine structuring element  & 0.531 & 0.524 & 0.523 & 0.537 \\
    \midrule
    \multicolumn{5}{@{}l}{\emph{$1/(1+\lambda_{\min})$ at tighter budgets}} \\
    40\% & 0.651 & 0.672 & 0.697 & 0.704 \\
    20\% & 0.597 & 0.614 & 0.664 & 0.663 \\
    \addlinespace[2pt]
    \emph{clutter fraction, 70\%} & 79.1\% & 75.1\% & 96.8\% & 95.7\% \\
    \emph{clutter fraction, 40\%} & 69.3\% & 67.3\% & 94.9\% & 94.3\% \\
    \emph{clutter fraction, 20\%} & 56.1\% & 55.0\% & 91.5\% & 91.1\% \\
    \bottomrule
  \end{tabular}
\end{table}

Gradient anisotropy is the only candidate of the five that clears chance by a margin worth reporting. It reaches $0.712$--$0.750$ against $0.52$--$0.65$ for four contrast-derived alternatives, and that set includes the classic local-contrast test and the product of the two strongest of them. It also gains the most on AI-TOD-v2, where the objects are far smaller while the clutter (coastlines, runways, field edges) is not. Discrimination then falls as the budget tightens, from $0.731$ to $0.614$ on VisDrone. We attribute that to the population and not the statistic: at a tight budget the clutter fraction among gate-visible nodes falls from $75\%$ to $55\%$, so there is progressively less clutter left to distinguish.

\paragraph{Scope of the audit.} The audit isolates the mechanism and the AP tables price it. It establishes that $\lambda_{\min}$ identifies the nodes we built the method to identify, on $150$ VisDrone val and $150$ AI-TOD-v2 val images, with no detector in the loop. Accuracy is a separate measurement, and we take it from AP under the stated protocol, on the full evaluation split. Together they are what a mechanism claim needs: the audit shows the statistic does what we say, and \cref{sec:results} shows what that buys.

\section{Pixel-score ablation}
\label{sec:suppl:scorers}

The choice of pixel score is not free. Eight training-free candidates held to one budget span $2.30$ AP$_{50}$, $2.39$ AP$_{75}$ and $2.34$ AP$_S$. \Cref{tab:suppl:scorers} prints the ranking behind the spread the main paper quotes. We hold everything but the statistic: eight patch scores at a fixed $70\%$ budget under Single-pass, no retraining on the dense checkpoint. The score that ranks patches is our only variable, and no training enters.

\begin{table}
    \caption{Eight patch scores on the metrics beyond AP. The ranking is the same on every metric we measured: against the AP order AP$_{50}$ transposes nothing, AP$_S$ transposes one adjacent pair by $0.02$ and AP$_{75}$ one by $0.14$, both inside the $0.15$ AP seed floor. Fixed $70\%$ budget, no retraining on the dense checkpoint under Single-pass; VisDrone val, $548$ images, maxDets $500$, with detector, budget and protocol identical across rows. Rows are in AP order and AP itself is not reprinted; the main paper quotes its spread in prose. Bold is the best value in a column. $^{\ast}$Upsample MSE is a valid measurement at this budget but cannot be budget-matched under a quadtree, for the reason below.}
  \label{tab:suppl:scorers}
  \centering
  \footnotesize
  \setlength{\tabcolsep}{5pt}
  \begin{tabular}{@{}lrrr@{}}
    \toprule
    Statistic & AP$_{50}$ & AP$_{75}$ & AP$_S$ \\
    \midrule
    $\lambda_{\min}$ (structure tensor) & \textbf{73.61} & \textbf{51.01} & \textbf{38.91} \\
    top-hat                             & 73.45 & 50.81 & 38.81 \\
    max deviation, Lab                  & 73.39 & 50.66 & 38.83 \\
    max deviation                       & 73.02 & 50.25 & 38.47 \\
    upsample MSE$^{\ast}$               & 72.75 & 49.73 & 38.06 \\
    gradient energy                     & 72.61 & 49.86 & 37.98 \\
    Laplacian                           & 72.23 & 49.20 & 37.43 \\
    entropy                             & 71.31 & 48.62 & 36.57 \\
    \midrule
    \emph{spread}                       & 2.30  & 2.39  & 2.34 \\
    \bottomrule
  \end{tabular}
\end{table}

\paragraph{Where $\lambda_{\min}$ pays.} Two measurements in this document constrain the design between them. In the ranking above, $\lambda_{\min}$ sits at the head of the eight candidate pixel scores, within the floor of the top-hat we ship. In \cref{tab:suppl:auroc} it separates clutter from object-bearing nodes at $0.712$--$0.750$ where every contrast score reaches at most $0.65$, so the information it carries is not the information a contrast score carries. The statistic is informative, and the gate is where it pays. We spend it once, at the gate, where it decides whether to stop rather than where to look.

\paragraph{Why the bake-off runs under Single-pass.} We run the bake-off under Single-pass because one of the eight candidates has no legal budget under a quadtree at all. Upsample MSE scores a patch by how much detail is lost when the region is rendered coarsely. Our implementation evaluates it at a scale factor of $1$ on the finest grid, where the downsample--upsample round trip is the identity, so the map is identically zero. Under Single-pass this is harmless. Each scale gets its own percentile, and a percentile of a constant array still selects exactly $1-p/100$ of the patches, so the budget is exact and the row above is a valid measurement. Under any quadtree it is fatal. A quadtree takes a single global threshold from the finest grid, and that threshold is zero at every percentile. The descent then stops depending on the percentile, and the retained fraction flatlines at $0.2493$ across the whole sweep. The scorer cannot be held to a stated budget under the splitter this paper uses, so it appears here and nowhere else. Its two coarser scales carry real signal, so no aggregate view of the scorer shows the defect. We found it by checking that retention must be monotone in the percentile that sets it, a test the budget-inversion protocol of \cref{sec:suppl:budget} makes routine and an absolute-threshold tokenizer never runs.

\section{Budget ladders without retraining}
\label{sec:suppl:ladders}

These ladders separate what the splitter buys from what the gate buys, at matched budgets and with no weight changed. The dashed curves of \cref{fig:pareto} are the two families: we swap each tokenizer into the dense checkpoint and change nothing else. \Cref{tab:suppl:treeladder} is the quadtree family. Single-pass is the other curve, and we give it in prose because its shape is one law and two endpoints. Twelve thresholds from $p=5$ to $p=80$, evaluated over the same $548$ VisDrone val images at maxDets $500$, retain $0.965$ down to $0.400$ of the dense grid. A fifth of the tokens can be dropped for $0.25$ AP. At $p=10$ it holds $0.928$ retained and scores $48.32$ AP and $39.55$ AP$_S$, just below the dense checkpoint it was cut from. It is still at $48.28$ with $0.851$ retained. It holds $47.54$ AP and $38.81$ AP$_S$ at $0.700$ retained, still above every published VisDrone-val entry, and reaches $32.96$ and $23.00$ at $0.400$, where a single global percentile has no depth left to reallocate. The two families are different splitters on different percentile scales, so the controlled comparison between them is vertical, at a fixed retained fraction, never horizontal across a threshold.

\begin{table}
    \caption{Gate effect without retraining at five matched budgets. The effect grows as the budget tightens, peaks at $+1.02$ AP and $+1.18$ AP$_S$ at $40\%$, then flattens to $+1.02$ and $+1.07$ at $20\%$. The quadtree splitter on the dense checkpoint. VisDrone val, $548$ images, maxDets $500$. We reach each budget by inverting that tokenizer's own measured retention table, so the two rows hold the same token count in every column.}
  \label{tab:suppl:treeladder}
  \centering
  \footnotesize
  \setlength{\tabcolsep}{4pt}
  \begin{tabular}{@{}lrrrrr@{}}
    \toprule
    Tokenizer & 80\% & 70\% & 60\% & 40\% & 20\% \\
    \midrule
    \multicolumn{6}{@{}l}{\emph{AP}} \\
    Top-hat ($\mathcal{T}$)        & 48.18 & 47.76 & 46.54 & 39.54 & 20.17 \\
    Top-hat ($\mathcal{T}$) $+\,g$ & \textbf{48.23} & \textbf{47.94} & \textbf{47.01} & \textbf{40.56} & \textbf{21.18} \\
    \emph{gate effect} & $+0.04$ & $+0.18$ & $+0.47$ & $+1.02$ & $+1.02$ \\
    \addlinespace[3pt]
    \multicolumn{6}{@{}l}{\emph{AP$_S$}} \\
    Top-hat ($\mathcal{T}$)        & 39.46 & 39.05 & 37.98 & 31.31 & 13.79 \\
    Top-hat ($\mathcal{T}$) $+\,g$ & \textbf{39.47} & \textbf{39.26} & \textbf{38.41} & \textbf{32.50} & \textbf{14.86} \\
    \emph{gate effect} & $+0.01$ & $+0.21$ & $+0.43$ & $+1.18$ & $+1.07$ \\
    \bottomrule
  \end{tabular}
\end{table}

The controlled comparison between the two families is vertical, at a fixed retained fraction. At $40\%$ the recursive descent is worth $+6.59$ AP and $+8.31$ AP$_S$ over the single-pass allocator: $39.54$ and $31.31$ against $32.96$ and $23.00$, under identical scores, identical budget, identical weights and identical protocol. This is the largest effect we measure anywhere in the paper, and it is what the descent buys before the gate does anything at all. \Cref{tab:suppl:treeladder} then isolates the gate on top of it.

\paragraph{An affine retention law, and why the quadtree needs a measured one.} Under Single-pass the retained fraction is an exact affine function of the threshold, $\text{retained}=1-0.0075\,p$. Measured over all $548$ val images the law reproduces every one of those twelve rungs to within $0.29$ percentage points, and to within $0.03$ percentage points at every rung from $p=40$ downward in budget. The reason is structural: the gate applies an independent percentile to each scale, and a percentile keeps exactly $1-p/100$ of the patches at each of them whatever the image contains. This law is Single-pass's alone, and its failure under a quadtree is why we built the budget protocol. A quadtree's token count depends on \emph{where} the busy nodes sit and on how deep the descent runs before it stops, so no closed form maps a threshold onto a budget. We therefore set every quadtree budget in this paper by inverting a measured retention table (\cref{sec:suppl:budget}), which is what lets us hold four tokenizers to $0.4001$ of the dense grid at once.

\section{Scale-resolved results on VisDrone}
\label{sec:suppl:bands}

The gate lifts the smallest objects most, and the size it favours is the mechanism's clearest prediction, so we open the bands here. \Cref{tab:suppl:bands40} gives all six COCO bands at the operating point.

\begin{table*}
    \caption{All six COCO bands at the $40\%$ operating point. In five of the six columns the gain clears each tokenizer's own $\sigma$ and runs $0.9\times$ to $6.8\times$ the propagated $\sigma_{\Delta}$. VisDrone val, $548$ images, maxDets $500$. Each mean is three independent $8$-epoch finetunes at a budget matched to $0.4001$ retained, with $\sigma$ their sample standard deviation. The two rows differ only in the gate $g$. On the difference row $\sigma_{\Delta}=\sqrt{\sigma_1^2+\sigma_2^2}$, the propagated spread of a single-run difference. AP$_M$ is the column where the gain does not clear $\sigma$; AP$_L$ covers $1068$ of $38{,}759$ scored instances.}
  \label{tab:suppl:bands40}
  \centering
  \footnotesize
  \setlength{\tabcolsep}{6pt}
  \begin{tabular}{@{}lrrrrrr@{}}
    \toprule
    Tokenizer & AP & AP$_{50}$ & AP$_{75}$ & AP$_S$ & AP$_M$ & AP$_L$ \\
    \midrule
    Top-hat ($\mathcal{T}$)        & 44.16 \tiny$\pm$0.08 & 69.47 \tiny$\pm$0.04 & 46.70 \tiny$\pm$0.20 & 35.44 \tiny$\pm$0.02 & 56.59 \tiny$\pm$0.13 & 69.67 \tiny$\pm$0.57 \\
    Top-hat ($\mathcal{T}$) $+\,g$ & \textbf{44.64} \tiny$\pm$0.10 & \textbf{70.01} \tiny$\pm$0.07 & \textbf{47.28} \tiny$\pm$0.14 & \textbf{36.03} \tiny$\pm$0.12 & 56.72 \tiny$\pm$0.22 & \textbf{71.01} \tiny$\pm$1.31 \\
    \addlinespace[2pt]
    \emph{gate effect}             & $+0.48$ \tiny$\pm$0.13 & $+0.54$ \tiny$\pm$0.08 & $+0.59$ \tiny$\pm$0.24 & $+0.59$ \tiny$\pm$0.12 & $+0.14$ \tiny$\pm$0.26 & $+1.34$ \tiny$\pm$1.43 \\
    \bottomrule
  \end{tabular}
\end{table*}

The gate moves the small-object band further than the aggregate at both tight budgets. On AP$_S$ it is worth $+0.59$ against $+0.48$ AP at $40\%$, and $+1.07$ against $+1.02$ at $20\%$ without retraining. This is the ordering the mechanism predicts. The gate declines to spend tokens on one-dimensional structure, and the descent re-spends them at the finest scale, where the small objects are. We take the ordering as evidence: a gate working by some other route would have no reason to favour AP$_S$ over AP. At $80\%$ the effect is $+0.01$ AP$_S$ without retraining, zero to the precision we measure, and that is what a stop rule is worth where tokens are not scarce.

\section{AI-TOD-v2 by area band and class}
\label{sec:suppl:aitod}

AI-TOD-v2 supports a decomposition VisDrone does not. Every AI-TOD-v2 evaluation in this paper writes all four of the benchmark's area bands and, separately, the very-tiny band broken out over all eight classes, where VisDrone's evaluations carry no per-class field. We use that here to test the gate against a prediction the mechanism makes about specific object geometries, a check the first dataset cannot run. The tokenizer and the rank $r=0.20$ are carried across unchanged. We open the bands in \cref{tab:suppl:aitodbands}, carrying the factorial the main paper states in prose, and the classes in \cref{tab:suppl:aitodclass}.

\begin{table}
    \caption{AI-TOD-v2 area bands, by budget. The gate lifts every band at the $70\%$ operating point we report on this benchmark; below it the very-tiny band alone turns over, the one place the ordering runs opposite to VisDrone, and every cell is printed in full. AI-TOD-v2 test, the $14{,}018$-image split, maxDets $1500$ at $2048$; AP$_{\mathrm{vt}}$ is very-tiny (under $8^2$ px), then tiny, small and medium, no large band here. Every row runs without retraining on one shared NWD dense checkpoint, budgets matched by inverting each tokenizer's own retention table (\cref{tab:suppl:inversion}). There is no $80\%$ block: at $r=0.20$ the gate alone caps retention at $0.785$ here.}
  \label{tab:suppl:aitodbands}
  \centering
  \footnotesize
  \setlength{\tabcolsep}{2.6pt}
  \begin{tabular}{@{}lrrrrrr@{}}
    \toprule
    Tokenizer & Budget & AP & AP$_{\mathrm{vt}}$ & AP$_{\mathrm{t}}$ & AP$_{\mathrm{s}}$ & AP$_{\mathrm{m}}$ \\
    \midrule
    VGTok, $100\%$ tokens & 100\% & 37.27 & 19.51 & 37.47 & 43.57 & 56.04 \\
    \midrule
    Top-hat ($\mathcal{T}$)                       & 70\% & 35.58 & 17.16 & 35.88 & 41.14 & 52.46 \\
    Top-hat ($\mathcal{T}$) $+\,g$                & 70\% & 35.86 & \textbf{18.58} & \textbf{36.23} & 41.47 & 52.85 \\
    Top-hat ($\mathcal{T}$) $\oplus\,\lambda$     & 70\% & 35.65 & 18.22 & 36.01 & 41.32 & 52.85 \\
    Top-hat ($\mathcal{T}$) $\oplus\,\lambda + g$ & 70\% & \textbf{35.89} & 18.55 & 36.20 & 41.47 & 52.83 \\
    \midrule
    Top-hat ($\mathcal{T}$)                       & 40\% & 30.17 & 14.83 & 30.18 & 34.56 & \textbf{45.10} \\
    Top-hat ($\mathcal{T}$) $+\,g$                & 40\% & 30.30 & 14.40 & 30.52 & \textbf{34.65} & 44.59 \\
    Top-hat ($\mathcal{T}$) $\oplus\,\lambda$     & 40\% & \textbf{30.40} & \textbf{14.98} & \textbf{30.70} & 34.64 & 45.00 \\
    Top-hat ($\mathcal{T}$) $\oplus\,\lambda + g$ & 40\% & 30.27 & 14.48 & 30.58 & 34.57 & 44.77 \\
    \midrule
    Top-hat ($\mathcal{T}$)                       & 20\% & 20.51 & \textbf{9.79} & 20.20 & 23.53 & \textbf{33.71} \\
    Top-hat ($\mathcal{T}$) $+\,g$                & 20\% & 20.82 & 9.45 & 20.58 & 23.84 & 33.63 \\
    Top-hat ($\mathcal{T}$) $\oplus\,\lambda$     & 20\% & \textbf{21.13} & 9.61 & \textbf{21.08} & \textbf{24.20} & 33.49 \\
    Top-hat ($\mathcal{T}$) $\oplus\,\lambda + g$ & 20\% & 20.64 & 8.92 & 20.40 & 23.86 & 32.91 \\
    \bottomrule
  \end{tabular}
\end{table}

Read down the $70\%$ block. The gate is worth $+0.29$ AP and moves every band the same way: $+1.42$ very-tiny, $+0.35$ tiny, $+0.33$ small, $+0.38$ medium. At $40\%$ it is worth $+0.13$ AP and the bands disagree, $-0.43$ very-tiny against $+0.34$ tiny. At $20\%$ it is worth $+0.31$ AP, still $+0.38$ on the tiny band and still negative on the very-tiny one. The largest band that keeps its sign is the tiny band, which holds $275{,}050$ of the split's $376{,}121$ instances. The band that flips holds the fewest instances, and the fewest classes carrying them, which is what we open in \cref{tab:suppl:aitodclass}.

\begin{table*}
    \caption{Per-class AP on the very-tiny band. The $+1.42$ AP$_{\mathrm{vt}}$ the gate is worth at $70\%$ is one class: airplane moves $+10.90$ on $17$ boxes. AI-TOD-v2 test, $14{,}018$ images, maxDets $1500$; both tokenizer rows run without retraining on the same NWD dense checkpoint at a budget matched to $70\%$ by inverting each tokenizer's own retention table, so the gate is the only difference between them. \emph{Very-tiny boxes} is that class's instance count in the band, counted from the test annotations. AP$_{\mathrm{vt}}$ is an unweighted mean over the eight classes, so each carries one eighth of it however few boxes it holds, and the \emph{share} row is that eighth.}
  \label{tab:suppl:aitodclass}
  \centering
  \footnotesize
  \setlength{\tabcolsep}{4pt}
  \begin{tabular}{@{}lrrrrrrrrr@{}}
    \toprule
    & vehicle & person & ship & storage-tank & bridge & swimming-pool & airplane & wind-mill & AP$_{\mathrm{vt}}$ \\
    \emph{very-tiny boxes} & 38{,}504 & 5{,}934 & 2{,}930 & 1{,}522 & 41 & 39 & 17 & 8 & 48{,}995 \\
    \midrule
    VGTok, $100\%$ tokens & 19.39 & 11.07 & 35.07 & 32.20 & 4.84 & 27.74 & 20.34 & 5.46 & 19.51 \\
    \midrule
    Top-hat ($\mathcal{T}$)        & 19.61 & 10.52 & 32.66 & 32.01 & \textbf{3.17} & 25.37 & 8.14 & \textbf{5.80} & 17.16 \\
    Top-hat ($\mathcal{T}$) $+\,g$ & 19.61 & \textbf{10.64} & \textbf{33.25} & \textbf{32.23} & 3.05 & \textbf{26.03} & \textbf{19.05} & 4.80 & \textbf{18.58} \\
    \addlinespace[2pt]
    \emph{gate effect} & $-0.01$ & $+0.13$ & $+0.59$ & $+0.21$ & $-0.12$ & $+0.66$ & $+10.90$ & $-1.00$ & $+1.42$ \\
    \emph{share of the mean} & $-0.001$ & $+0.016$ & $+0.073$ & $+0.027$ & $-0.014$ & $+0.082$ & $+1.363$ & $-0.126$ & $+1.419$ \\
    \bottomrule
  \end{tabular}
\end{table*}

The gate moves five of the eight classes up, and three of the four classes holding more than a thousand very-tiny boxes move up: $-0.01$, $+0.13$, $+0.21$ and $+0.59$ on vehicle, person, storage-tank and ship. None of those turns on a handful of detections. The aggregate is a different quantity. AP$_{\mathrm{vt}}$ on this benchmark is an unweighted mean over eight classes, and four of them hold $105$ of the $48{,}995$ very-tiny boxes between them. Half of the metric therefore rests on a fifth of a percent of the instances. Airplane alone contributes $+1.363$ of the $+1.42$ and the other seven classes net to $+0.056$. We measured that column under a real control (same checkpoint, same budget, same protocol) on $17$ boxes. We read the direction from the four populous classes.

The gate's one large per-class loss lands where the mechanism says it must. The gate declines to split one-dimensional structure, so an object that \emph{is} one-dimensional should be the object it damages. Wind-mill, a hub with three long thin blades seen from nadir, falls $1.00$, from $5.80$ to $4.80$. That is the largest per-class loss we record anywhere, on the one class in this benchmark whose geometry the gate is built to reject. We read it as evidence the gate acts through the mechanism we claim, not through some correlate of it. A statistic working by another route would have no reason to single out this class. Swimming-pool is the opposite geometry (a compact rectangle of near-uniform texture) and rises $0.66$ on $39$ boxes, inside anything we would call resolution on this metric.

The prediction is directional, and the class that carries it is wind-mill; bridge, the other one-dimensional class here, moves $-0.12$ and is the hardest class on the split for every tokenizer, including VGTok at $100\%$ tokens, $4.84$ AP$_{\mathrm{vt}}$ there against $3.17$ ungated.

\section{Sensitivity to the gate rank $r$}
\label{sec:suppl:rank}

VGTok adds exactly one hyperparameter, the gate rank $r$, and we ship it at $r=0.20$ on both datasets without retuning. \Cref{tab:suppl:rank} is our accuracy sweep behind that value: $0.20$ is the argmax at every budget, and AP falls on both sides of it, so the shipped setting is an interior optimum. \Cref{tab:suppl:inversion} separately shows what $r$ does to the budget, which is a different question.

\begin{table}
    \caption{Rank sweep at three matched budgets. The shipped $r=0.20$ is the argmax of AP in all three columns and AP falls on both sides of it, so this is an interior optimum rather than an untested default. Twelve evaluations without retraining of one dense checkpoint, tokenizer swapped in and no weight changed: VisDrone val, $548$ images, maxDets $500$, native $2048\times1152$, splitter and scorer fixed so that $r$ is the only variable. We match budgets per rank by inverting each rank's own retention table, so the percentiles \cref{tab:suppl:inversion} prints differ by row. Bold is the best value in a column among the four comparable rows. No training seed enters an evaluation without retraining, so these differences are exact rather than draws against the seed floor. A fifth rung at $r=0.40$ falls further, to $40.17$ and $20.79$. The rank was chosen on VisDrone val, then carried onto AI-TOD-v2 and never re-tuned.}
  \label{tab:suppl:rank}
  \centering
  \footnotesize
  \setlength{\tabcolsep}{6pt}
  \begin{tabular}{@{}lrrr@{}}
    \toprule
    & \multicolumn{3}{c}{AP, at a matched token budget} \\
    \cmidrule(lr){2-4}
    Gate rank $r$ & 70\% & 40\% & 20\% \\
    \midrule
    0.05                  & 47.89 & 40.19 & 20.80 \\
    0.10                  & 47.91 & 40.49 & 21.13 \\
    0.20 \emph{(shipped)} & \textbf{47.94} & \textbf{40.56} & \textbf{21.18} \\
    0.30                  & 47.91 & 40.21 & 20.92 \\
    \bottomrule
  \end{tabular}
\end{table}

The peak is real and sits at $r=0.20$ at every budget, and it sits on a plateau; both halves of that are worth stating. Moving out to $r=0.05$ costs $0.37$ AP at the $40\%$ budget and $0.38$ at $20\%$; moving out to $r=0.30$ costs $0.35$ and $0.26$, so a rank away from $0.20$ in either direction is measurably worse. Inside the plateau the margin over the neighbouring rank $r=0.10$ is $0.03$, $0.07$ and $0.05$ AP. The table settles that the answer lies in that neighbourhood, and the flatness there is what makes the knob transferable: any $r$ in it costs at most $0.07$ AP, so one setting carries across datasets. We then put that to the test by carrying $r=0.20$ onto AI-TOD-v2 unchanged.

\Cref{tab:suppl:inversion} explains the shape at the loose end. At the $70\%$ budget the four ranks already spend most of their reduction on different mechanisms. Rank $r=0.05$ reaches that budget at $p\,47.12$ and $r=0.30$ at $p\,8.66$, where the threshold has almost stopped contributing. The four rows are therefore increasingly unlike one another even though they hold the same token count, and the AP they reach is within $0.05$ of a common value. At the tight end the threshold does most of the work in every row ($p\,91$--$p\,94$ at the $20\%$ budget), and the rank becomes a genuine second variable on top of it. That is where the $0.38$ AP spread appears.

\section{Convergence of the dense reference}
\label{sec:suppl:dense}

We make every comparison against VGTok at $100\%$ of tokens, $48.38$ AP, so that ceiling has to be converged for the margins beneath it to mean anything. We check it twice, by running longer and by restarting the schedule, and it holds both times.

\begin{figure}[t]
  \centering
  \includegraphics[width=\columnwidth]{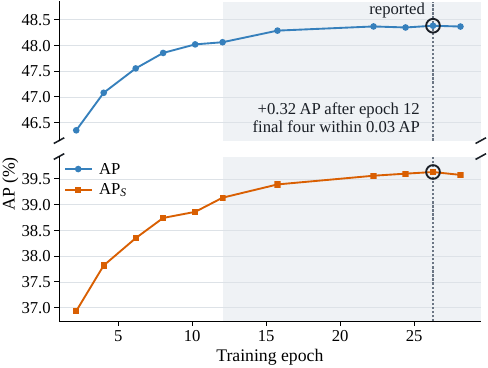}
    \caption{Dense epoch ladder, eleven checkpoints of one training run. The ladder saturates: everything after epoch $12$ is worth $+0.32$ AP in total, and the last twelve epochs of that only $+0.10$. Every point is scored under the reported protocol: VisDrone val, $548$ images, maxDets $500$, ignore regions at IoF $\ge 0.5$, native $2048\times1152$. The ringed point is the checkpoint we report and the maximum of both curves.}
  \label{fig:suppl:dense}
\end{figure}

The ladder saturates. The four points from epoch $22$ to epoch $28$ lie inside $0.03$ AP of one another, and most of the curve is spent by epoch $12$. We read the annealing check as the sharper of the two, because a saturated cosine schedule is where one expects a fresh anneal to find another few tenths. It finds none, which is what makes the dense reference a fair one. Restarting from the epoch-$28$ point with a fresh schedule dips to $47.70$, peaks at $48.30$ and finishes at $48.15$, all at or under the checkpoint we report. The reference is a finished schedule, and the headroom we claim is not a training artifact. One figure moves: AP$_S$ at the link's best point reaches $39.66$ against $39.63$, and we print it.

\section{Seed replication and the seed floor}
\label{sec:suppl:seeds}

Every trained VisDrone number in this paper is a three-seed mean. That is the $2$ trained cells of \cref{tab:gate}, $6$ full $8$-epoch finetunes, and it is what lets us separate the gate effect from seed noise instead of asserting it. We list the $6$ finetunes at the operating point individually in \cref{tab:suppl:seeds}, on AP and on AP$_S$, with the sample standard deviation beside each mean. The runs differ only in data order and initialisation seed, and we score each under the VisDrone protocol at maxDets $500$.
\begin{table}
    \caption{Seed replicates at the $40\%$ budget. The gated and ungated tokenizers do not overlap on either metric: worst gated $44.55$ against best ungated $44.25$ on AP, and $35.89$ against $35.46$ on the noisier AP$_S$, so the reported gate effect is not seed noise. VisDrone2019-DET val, $548$ images, maxDets $500$; all six runs sit at one budget, $0.4001$ retained (\cref{tab:suppl:budget}), and the two blocks score them on AP and AP$_S$. \emph{Run 1} is the original cell and the rest independent repeats; $\sigma$ is their sample standard deviation. We pool both seed floors over the factorial instead of reading them off one cell: within-cell $\sigma$ over the $2$ trained VisDrone cells is $0.0935$ AP and $0.0880$ AP$_S$, so a difference of two three-seed means has standard error $0.0763$ and $0.0718$; twice that, $0.15$ AP and $0.14$ AP$_S$, is the smallest difference we argue for.}
  \label{tab:suppl:seeds}
  \label{tab:suppl:seedsaps}%
  \centering
  \footnotesize
  \setlength{\tabcolsep}{5pt}
  \begin{tabular}{@{}lrrrrr@{}}
    \toprule
    Tokenizer (AP) & Run 1 & Run 2 & Run 3 & Mean & $\sigma$ \\
    \midrule
    Top-hat ($\mathcal{T}$)                       & 44.10 & 44.25 & 44.13 & 44.16 & 0.08 \\
    Top-hat ($\mathcal{T}$) $+\,g$                & 44.75 & 44.63 & 44.55 & \textbf{44.64} & 0.10 \\
    \midrule
    Tokenizer (AP$_S$) & Run 1 & Run 2 & Run 3 & Mean & $\sigma$ \\
    \midrule
    Top-hat ($\mathcal{T}$)                       & 35.42 & 35.46 & 35.45 & 35.44 & 0.02 \\
    Top-hat ($\mathcal{T}$) $+\,g$                & 36.12 & 35.89 & 36.08 & \textbf{36.03} & 0.12 \\
    \bottomrule
  \end{tabular}
\end{table}

The spread is not constant across tokenizers. Over the $2$ trained VisDrone cells behind this paper the range of three finetunes runs from $0.155$ to $0.205$ AP and from $0.043$ to $0.229$ AP$_S$. We therefore quote one floor everywhere, pooled over the whole factorial: the alternative, taking the widest observed range, is not a standard error, and its expected value climbs with the number of cells inspected, so replicating more of the grid would have raised our own bar. The $40\%$ separation then holds on AP$_S$ as well as AP, the metric we aim at and the noisier of the two.

\section{Token-budget matching protocol}
\label{sec:suppl:budget}

VGTok makes the retained fraction an operator input rather than an outcome, and this section is the audit that shows it holds. Under a quadtree the token count depends on where the busy nodes sit, and no closed form maps a threshold onto a budget. We therefore set every threshold by inverting a measured retention table, then re-measure the result on the split we report. All four tokenizers land on $0.4001$ of the dense grid. \Cref{tab:suppl:budget} is our re-measurement and \cref{tab:suppl:inversion} is the table our thresholds came out of.

\begin{table}
    \caption{Token-budget audit of the $40\%$ column. All four tokenizers land on $0.4001$ of the dense grid, so the gate's margin at this budget is measured at matched tokens. We measure the retained fraction over all $548$ VisDrone val images at $2048\times1152$ against a dense grid of $9216$ tokens, for each tokenizer at the exact threshold we deployed it at; \emph{Tokens} is the per-image mean.}
  \label{tab:suppl:budget}
  \centering
  \footnotesize
  \setlength{\tabcolsep}{5pt}
  \begin{tabular}{@{}llrr@{}}
    \toprule
    Tokenizer & Threshold & Tokens & Retained \\
    \midrule
    Top-hat ($\mathcal{T}$)                       & $p\,81.88$ & 3687.9 & 0.4001 \\
    Top-hat ($\mathcal{T}$) $+\,g$                & $p\,74.82$ & 3687.3 & 0.4001 \\
    Top-hat ($\mathcal{T}$) $\oplus\,\lambda$     & $p\,81.96$ & 3687.6 & 0.4001 \\
    Top-hat ($\mathcal{T}$) $\oplus\,\lambda + g$ & $p\,74.70$ & 3688.1 & 0.4001 \\
    \bottomrule
  \end{tabular}
\end{table}

\begin{table}
    \caption{Inverted retention tables, five budgets by five gate ranks. Rank and threshold are not interchangeable: raising $r$ shifts the whole table left until the loosest budgets fall off it, and by $r=0.30$ the $80\%$ budget is unreachable on both datasets. Each cell is the percentile that delivers the column's budget at that rank, from a $32$-rung measured sweep over all $548$ VisDrone val images and $2804$ AI-TOD-v2 images. The italic row is the shipped setting, $r=0.20$, used unchanged on both. \emph{Unreachable} is structural rather than unrun: at that rank the gate alone caps retention below the budget, which is why \cref{tab:suppl:aitodbands} has no $80\%$ block. We invert percentiles on the AI-TOD-v2 \emph{val} split and verify them against the $14{,}018$-image \emph{test} split both tokenizers are scored on, where each overshoots its target by at most $0.15$ percentage points.}
  \label{tab:suppl:inversion}
  \centering
  \scriptsize
  \setlength{\tabcolsep}{3pt}
  \begin{tabular}{@{}lrrrrr@{}}
    \toprule
    Gate rank $r$ & 80\% & 70\% & 60\% & 40\% & 20\% \\
    \midrule
    \multicolumn{6}{@{}l}{\emph{VisDrone val}, $548$ images} \\
    none \emph{(= $\mathcal{T}$)} & 37.91 & 51.53 & 63.44 & 81.88 & 94.58 \\
    0.05                    & 32.16 & 47.12 & 60.28 & 80.45 & 94.21 \\
    0.10                    & 25.56 & 42.03 & 56.54 & 78.82 & 93.79 \\
    \emph{0.20 (shipped)}   & \emph{5.62} & \emph{29.06} & \emph{47.01} & \emph{74.82} & \emph{92.76} \\
    0.30                    & \emph{unreachable} & 8.66  & 33.43 & 69.16 & 91.38 \\
    0.40                    & \emph{unreachable} & \emph{unreachable} & 12.08 & 60.66 & 89.41 \\
    \addlinespace[3pt]
    \multicolumn{6}{@{}l}{\emph{AI-TOD-v2}, $2804$ images} \\
    none \emph{(= $\mathcal{T}$)} & 46.79 & 59.79 & 69.98 & 84.90 & 95.17 \\
    0.10                    & 32.11 & 50.98 & 64.13 & 82.52 & 94.51 \\
    \emph{0.20 (shipped)}   & \emph{unreachable} & \emph{36.70} & \emph{55.70} & \emph{79.29} & \emph{93.66} \\
    0.30                    & \emph{unreachable} & \emph{unreachable} & 41.96 & 74.69 & 92.53 \\
    \bottomrule
  \end{tabular}
\end{table}

The audit is exact. All four tokenizers realise $0.4001$ of the dense grid at the thresholds we deployed them at, so the gate's margin at $40\%$ is measured at matched tokens rather than inferred across a budget gap.

\Cref{tab:suppl:inversion} answers the question a reader asks on seeing two knobs. At the $40\%$ budget we run the shipped setting at $p\,74.82$, where the ungated control needs $p\,81.88$ to buy the same tokens. At $r=0.40$ we reach that budget at $p\,60.66$. At the loosest budget the two knobs collide. The shipped tokenizer reaches $80\%$ only at $p\,5.62$, so the gate does nearly all of the reduction and the threshold almost none. That is the regime where the gate is worth $+0.04$ AP, level with the ungated control against a $0.15$ floor, and the reason is mechanical: with tokens this plentiful, there is no tightening trade-off for a stop rule to exploit.

\paragraph{The AI-TOD-v2 budgets are verified on the split they are scored on.} We invert each retention table on the $2804$-image AI-TOD-v2 \emph{val} split, then check the percentiles it produces against the $14{,}018$-image \emph{test} split that every AI-TOD-v2 AP here is scored on. Both tokenizers overshoot, and both overshoot by the same small amount. The ungated control realises $0.8015$, $0.7012$, $0.6011$, $0.4015$ and $0.2010$ against targets of $0.80$ down to $0.20$; the shipped gated tokenizer realises $0.7011$, $0.6009$ and $0.4006$ against $0.70$, $0.60$ and $0.40$. Every error is positive and none exceeds $0.15$ percentage points, so this is a common bias between two splits and not scatter. A bias shared by both tokenizers cancels out of the difference between them. No AI-TOD-v2 comparison in this paper rests on a budget match looser than a tenth of a percentage point.

\section{Sensitivity to input resolution}
\label{sec:suppl:res}

The cost of pruning falls as the input grows, which is the direction high-resolution detection is moving. We fix the protocol at the native $2048\times1152$ throughout the paper, and \cref{tab:suppl:res} shows what happens if it is not, for VGTok at $100\%$ and at $70\%$ of tokens, on the same checkpoint and the same split.

\begin{table}
    \caption{VGTok at two budgets across four input resolutions. The cost of pruning falls as the input grows: the gap between full and $70\%$ retention narrows monotonically from $1.40$ AP at $1536$ to $0.64$ at $2560$, so the tokenizer gets cheaper in accuracy terms exactly where high-resolution detection is headed. Same checkpoint, same $548$-image VisDrone val split, maxDets $500$; only the input resolution changes. The $70\%$ column is Single-pass evaluated without retraining, the tokenizer we measured at all four resolutions, rather than the shipped quadtree one; the reading is about how the two columns move together at a fixed retention and does not depend on which tokenizer fills the second pair. \emph{Gap} is $100\%$ minus $70\%$ on AP, taken on unrounded values: at $2048$ it is $0.85$ where the printed cells differ by $0.84$.}
  \label{tab:suppl:res}
  \centering
  \footnotesize
  \setlength{\tabcolsep}{4pt}
  \begin{tabular}{@{}lrrrrr@{}}
    \toprule
    & \multicolumn{2}{c}{VGTok, $100\%$} & \multicolumn{2}{c}{VGTok, $70\%$} & \\
    \cmidrule(lr){2-3}\cmidrule(lr){4-5}
    Evaluation resolution & AP & AP$_S$ & AP & AP$_S$ & \emph{gap}, AP \\
    \midrule
    1536                            & 45.52 & 35.97 & 44.12 & 34.75 & $+1.40$ \\
    2048 \emph{(the protocol)}      & 48.38 & 39.63 & 47.54 & 38.81 & $+0.85$ \\
    2304                            & 48.86 & 40.39 & 48.11 & 39.73 & $+0.75$ \\
    2560                            & 49.39 & 41.04 & 48.75 & 40.37 & $+0.64$ \\
    \bottomrule
  \end{tabular}
\end{table}

Going from $2048$ to $2560$ is worth $+1.01$ AP at $100\%$ of tokens and $+1.21$ at $70\%$, and going down to $1536$ costs $-2.86$ and $-3.41$. The two columns move together, and the pruned one gains more at each step up, which is why the gap closes. We print the off-protocol rows in full, including the $49.39$ AP at $2560$ that sits above the number we claim. The protocol is fixed at $2048$ throughout, and every number we report is measured there.

\section{What the \texorpdfstring{$\lambda_{\min}$}{lambda-min} gate declines to refine}
\label{sec:suppl:failure}

The gate fires on the top $r$ of a ranking, and \cref{sec:suppl:auroc} measures that ranking: at an AUROC of $0.712$--$0.750$, the best of the five candidates by a wide margin, it orders a randomly drawn clutter/object pair correctly between $71\%$ and $75\%$ of the time. Ranking quality is not itself a cost, and this section prices the cost. \Cref{tab:suppl:failure} turns it into the quantity we care about: not how many nodes we mis-rank, but how much of what the gate declines to spend was describing an object. A gated node is still emitted, at its own coarse scale, so a finest-grid token counted here is a region described coarsely rather than a region dropped. Two effects set that rate, and the numbers separate them.

\begin{table}
    \caption{Coarsened tokens, by distance to the nearest object. The rate follows the budget, not the statistic: at the $40\%$ operating point $14.6\%$ of the tokens the gate declines to spend lie on an object, and on AI-TOD-v2 the figure is $1.4\%$ or below at every budget we report. The last four columns bin every finest-grid token the gate declines to spend by distance from its node to the nearest ground-truth box; the region is still described, at the gate's own coarse scale. Shipped statistic at rank $r=0.20$, on $150$ VisDrone \emph{val} and $150$ AI-TOD-v2 \emph{val} images. \emph{Fires on} is the fraction of gate-visible nodes that fire; \emph{frees} is the percentage points of the dense grid regained: at $20\%$, firing on $0.9\%$ of nodes frees $2.8$ points. Rows are rounded independently and need not sum to $100$.}
  \label{tab:suppl:failure}
  \centering
  \footnotesize
  \setlength{\tabcolsep}{3.4pt}
  \begin{tabular}{@{}lrrrrrr@{}}
    \toprule
    & & & \multicolumn{4}{c}{coarsened tokens, by distance} \\
    \cmidrule(lr){4-7}
    Budget & fires on & frees & on an object & 1 node & 2--3 nodes & farther \\
    \midrule
    \multicolumn{7}{@{}l}{\emph{VisDrone val}, $150$ images} \\
    70\% & 4.4\% & 13.1\,pp & 6.3\%  & 5.1\% & 12.3\% & 76.4\% \\
    40\% & 2.3\% & 6.9\,pp  & 14.6\% & 6.0\% & 9.9\%  & 69.5\% \\
    20\% & 0.9\% & 2.8\,pp  & 29.0\% & 6.4\% & 7.6\%  & 57.0\% \\
    \addlinespace[3pt]
    \multicolumn{7}{@{}l}{\emph{AI-TOD-v2 val}, $150$ images} \\
    70\% & 4.1\% & 12.2\,pp & 0.4\%  & 0.9\% & 3.3\%  & 95.4\% \\
    40\% & 2.2\% & 6.6\,pp  & 1.4\%  & 1.3\% & 3.3\%  & 94.1\% \\
    20\% & 0.9\% & 2.7\,pp  & 3.0\%  & 1.8\% & 3.0\%  & 92.1\% \\
    \bottomrule
  \end{tabular}
\end{table}

\paragraph{The population changes, not the gate.} On VisDrone we hold the statistic and the rank fixed. At $70\%$ of the budget $6.3\%$ of the tokens the gate declines to spend lie on an object, and $14.6\%$ at the $40\%$ operating point; at $20\%$, a budget below the range we report, $29.0\%$ do. Nothing about the gate changed between those rows; the population did. At a tight budget the nodes the descent still reaches are increasingly object-bearing, and the clutter fraction falls from $79\%$ to $56\%$ (\cref{tab:suppl:aurocbudget}). Firing on the top $r$ of a population that is mostly objects necessarily hits objects, and the gate is still worth $+1.02$ AP there. It need not be right every time to pay: the tokens it frees are re-spent where the descent needs them most, and the regions it coarsens are still described.

\paragraph{Compact two-dimensional clutter reads like an object.} $\lambda_{\min}$ is large wherever the gradient directions inside a window disagree. That is true of a vehicle. It is equally true of a roundabout, a tree crown, a rooftop vent cluster or a stack of pallets. $\lambda_{\min}$ measures the property a compact object and compact clutter share, and separating those two is outside what one second-moment eigenvalue can express; that is the ceiling holding the AUROC at $0.75$ rather than higher. What it costs is a saving and not an object: the gate keeps descending into clutter it could have declined, so those tokens are spent rather than freed, and the $+0.48$ AP the gate is worth at the operating point is measured with that ceiling already in it.

\section{Dense detector by input size}
\label{sec:suppl2:dense}

The dense rows of \cref{tab:equal} use the detector's default attention. \Cref{tab:suppl2:dense} repeats them on the packed attention that VGTok uses, by passing the dense checkpoint through the ungated quadtree at $p=0$, which keeps at least $99.5\%$ of the tokens; at the same input size the packed attention is $8$ to $11\%$ faster. \Cref{tab:suppl2:intervals} gives the differences between VGTok and both sets of dense rows at equal GFLOPs and at equal latency.

\begin{table*}
  \caption{The dense detector on the packed attention that VGTok uses and on its default attention, at the same input size and with no retraining. \emph{Input size} is the long side of the image. The packed rows pass the dense checkpoint through the ungated quadtree at $p=0$, and \emph{Tokens} is the percentage of tokens they keep. GFLOPs and median latency are end to end, as in \cref{tab:compute}. \emph{AP difference} is packed minus default, with its $95\%$ paired bootstrap interval (\cref{sec:suppl2:intervals}).}
  \label{tab:suppl2:dense}
  \centering
  \scriptsize
  \setlength{\tabcolsep}{4pt}
  \begin{tabular}{@{}lrrrrrrrrr@{}}
    \toprule
    & \multicolumn{5}{c}{Packed attention} & \multicolumn{3}{c}{Default attention} & \\
    \cmidrule(lr){2-6}\cmidrule(lr){7-9}
    Input size & Tokens, \% & GFLOPs & Latency, ms & AP & AP$_S$ & GFLOPs & Latency, ms & AP & AP difference \\
    \midrule
    \multicolumn{10}{@{}l}{\emph{VisDrone val, $548$ images}} \\
    1664 px & 99.6 & 8{,}120 & 122.1 & 46.47 & 37.18 & 8{,}645 & 133.5 & 46.50 & $-0.03$ [$-0.09$, $+0.05$] \\
    1792 px & 99.6 & 9{,}487 & 133.9 & 47.28 & 38.05 & 9{,}838 & 145.8 & 47.33 & $-0.05$ [$-0.09$, $+0.02$] \\
    1920 px & 99.6 & 11{,}013 & 148.0 & 47.88 & 39.05 & 11{,}160 & 160.8 & 47.86 & $+0.03$ [$-0.01$, $+0.07$] \\
    2048 px & 99.5 & 12{,}670 & 163.6 & 48.36 & 39.58 & 12{,}956 & 179.8 & 48.38 & $-0.02$ [$-0.07$, $+0.01$] \\
    \multicolumn{10}{@{}l}{\emph{VisDrone test-dev, $1{,}610$ images}} \\
    1792 px & 99.9 & 9{,}419 & 133.7 & 36.64 & 25.08 & 9{,}652 & 146.8 & 36.61 & $+0.03$ [0.00, $+0.08$] \\
    1920 px & 99.9 & 10{,}892 & 148.1 & 37.07 & 25.73 & 11{,}127 & 161.9 & 37.08 & $-0.01$ [$-0.04$, $+0.03$] \\
    2048 px & 99.9 & 12{,}937 & 170.1 & 37.62 & 26.43 & 13{,}455 & 191.7 & 37.61 & $+0.01$ [$-0.02$, $+0.05$] \\
    \bottomrule
  \end{tabular}
\end{table*}

\section{Token merging}
\label{sec:suppl2:tome}

We implemented ToMe~\cite{bolya2023tome} for this backbone and run it on the dense checkpoint with no retraining: in each block, between attention and the MLP, every full $24\times24$-token attention window loses $m$ tokens by bipartite matching of similar tokens, merging stays within a window, and after the last block every position takes the feature of the token it was merged into, so the neck and head see the full grid. \Cref{tab:suppl2:tome} gives every merge count we ran; $m$ is constant across blocks except in the last row, where it falls linearly from $2m$ in the first block to $0$ in the last. With $m=0$ the implementation reproduces the dense detector, $48.38$ AP.

\begin{table}
  \caption{ToMe on the dense checkpoint with no retraining, VisDrone val, $548$ images, maxDets $500$. $m$ is the number of tokens a full window loses in each block. \emph{Tokens} is the mean over the $24$ blocks of the tokens that attention and the MLP process, and \emph{final} the share left after the last block, both as a percentage of the dense grid. GFLOPs and median latency are end to end, as in \cref{tab:compute}.}
  \label{tab:suppl2:tome}
  \centering
  \footnotesize
  \setlength{\tabcolsep}{3.5pt}
  \begin{tabular}{@{}rlrrrrr@{}}
    \toprule
    $m$ & Schedule & Tokens, \% & Final, \% & AP & GFLOPs & Latency, ms \\
    \midrule
    0 & constant & 100.0 & 100.0 & 48.38 & 12{,}956 & 181.4 \\
    5 & constant & 89.5 & 79.4 & 47.77 & 11{,}773 & 193.6 \\
    10 & constant & 79.3 & 59.5 & 45.38 & 10{,}749 & 183.0 \\
    14 & constant & 70.7 & 42.6 & 42.25 & 9{,}948 & 175.7 \\
    19 & constant & 60.6 & 22.7 & 35.13 & 9{,}071 & 168.1 \\
    21 & decreasing & 41.1 & 12.5 & 20.17 & 7{,}496 & 155.5 \\
    \bottomrule
  \end{tabular}
\end{table}

\section{Pixel scores in the gated quadtree}
\label{sec:suppl2:scores}

\Cref{tab:suppl2:scores} repeats the comparison of \cref{tab:suppl:scorers} inside the gated quadtree of $\mathcal{T}+g$, with each score in place of the top-hat, at four budgets and with no retraining. It also gives the blended score $\oplus\,\lambda$, with and without the gate. Upsample MSE has no budget under a quadtree, for the reason given in \cref{sec:suppl:scorers}.

\begin{table}
  \caption{Pixel scores in the quadtree with no retraining, VisDrone val, $548$ images, maxDets $500$. The gated rows use the gate at $r=0.20$, as $\mathcal{T}+g$ does. $\oplus\,\lambda$ blends $\lambda_{\min}$ into the top-hat score, and \emph{blend effect} is the blended row minus the top-hat row with the same gate setting, taken on unrounded values. \emph{Unreachable} marks a budget no threshold reaches with that score.}
  \label{tab:suppl2:scores}
  \centering
  \scriptsize
  \setlength{\tabcolsep}{3pt}
  \begin{tabular}{@{}lrrrr@{}}
    \toprule
    Pixel score & 80\% & 70\% & 60\% & 40\% \\
    \midrule
    \multicolumn{5}{@{}l}{\emph{AP, gated quadtree}} \\
    top-hat \emph{(= $\mathcal{T}+g$)} & 48.23 & 47.94 & 47.01 & 40.56 \\
    $\lambda_{\min}$ (structure tensor) & 48.22 & 47.95 & 47.03 & 39.58 \\
    max deviation, Lab & 48.23 & 47.92 & 46.95 & 40.59 \\
    max deviation & 48.22 & 47.92 & 47.02 & 40.01 \\
    gradient energy & 48.23 & 47.92 & 46.83 & 38.24 \\
    Laplacian & 48.22 & 47.88 & 46.34 & 36.09 \\
    entropy & 48.22 & 47.73 & 45.92 & 36.04 \\
    upsample MSE & \emph{unreachable} & \emph{unreachable} & \emph{unreachable} & \emph{unreachable} \\
    top-hat $\oplus\,\lambda$ & 48.22 & 47.92 & 47.01 & 40.23 \\
    \emph{blend effect} & $-0.01$ & $-0.02$ & 0.00 & $-0.33$ \\
    \addlinespace[3pt]
    \multicolumn{5}{@{}l}{\emph{AP, ungated quadtree}} \\
    top-hat \emph{(= $\mathcal{T}$)} & 48.18 & 47.76 & 46.54 & 39.54 \\
    top-hat $\oplus\,\lambda$ & 48.21 & 47.79 & 46.62 & 40.24 \\
    \emph{blend effect} & $+0.02$ & $+0.03$ & $+0.08$ & $+0.70$ \\
    \midrule
    \multicolumn{5}{@{}l}{\emph{AP$_S$, gated quadtree}} \\
    top-hat \emph{(= $\mathcal{T}+g$)} & 39.47 & 39.26 & 38.41 & 32.50 \\
    $\lambda_{\min}$ (structure tensor) & 39.47 & 39.26 & 38.42 & 31.53 \\
    max deviation, Lab & 39.49 & 39.21 & 38.25 & 32.34 \\
    max deviation & 39.47 & 39.24 & 38.36 & 31.56 \\
    gradient energy & 39.46 & 39.28 & 38.22 & 30.01 \\
    Laplacian & 39.46 & 39.23 & 37.81 & 28.36 \\
    entropy & 39.48 & 39.02 & 37.17 & 27.30 \\
    upsample MSE & \emph{unreachable} & \emph{unreachable} & \emph{unreachable} & \emph{unreachable} \\
    top-hat $\oplus\,\lambda$ & 39.45 & 39.21 & 38.35 & 31.92 \\
    \emph{blend effect} & $-0.02$ & $-0.05$ & $-0.05$ & $-0.58$ \\
    \addlinespace[3pt]
    \multicolumn{5}{@{}l}{\emph{AP$_S$, ungated quadtree}} \\
    top-hat \emph{(= $\mathcal{T}$)} & 39.46 & 39.05 & 37.98 & 31.31 \\
    top-hat $\oplus\,\lambda$ & 39.47 & 39.16 & 37.94 & 32.13 \\
    \emph{blend effect} & $+0.01$ & $+0.10$ & $-0.04$ & $+0.82$ \\
    \bottomrule
  \end{tabular}
\end{table}

\section{Tokenizer stage times}
\label{sec:suppl2:stages}

\Cref{tab:suppl2:stages} splits one forward pass into stages, with the GPU synchronised after each stage. The tokenizer takes $12.9$ to $14.9$\,ms per image.

\begin{table*}
  \caption{Median stage times in ms per image, VisDrone val, $548$ images, one H100, fp16, no retraining. \emph{Dense} is the dense detector on its default attention at $2048$\,px; the last two columns are the dense detector on the packed attention that VGTok uses, from a second run. The GPU is synchronised after each stage, so the synchronised end-to-end time exceeds that of a normal, unsynchronised pass. $^{\ast}$Upper bounds, since each of ToMe's merge and RoPE calls is synchronised. \emph{Blocks and rest} is the time in the transformer blocks plus any time not measured separately.}
  \label{tab:suppl2:stages}
  \centering
  \footnotesize
  \setlength{\tabcolsep}{2.5pt}
  \begin{tabular}{@{}lrrrrrrrr@{}}
    \toprule
    & & \multicolumn{4}{c}{VGTok} & & \multicolumn{2}{c}{Dense, packed attention} \\
    \cmidrule(lr){3-6}\cmidrule(l){8-9}
    Stage & Dense & 80\% & 70\% & 60\% & 40\% & ToMe $m{=}10$ & 2048\,px & 1920\,px \\
    \midrule
    Top-hat score &  & 4.2 & 4.2 & 4.2 & 4.2 &  & 4.2 & 3.8 \\
    $\lambda_{\min}$ &  & 2.6 & 3.2 & 3.8 & 4.4 &  &  &  \\
    Quadtree and gate &  & 1.3 & 1.3 & 1.2 & 1.1 &  & 0.6 & 0.6 \\
    Packing &  & 3.9 & 4.1 & 4.2 & 4.3 &  & 3.0 & 3.0 \\
    Scatter &  & 0.9 & 0.9 & 0.8 & 0.8 &  & 0.5 & 0.5 \\
    \addlinespace[2pt]
    Tokenizer total &  & 12.9 & 13.7 & 14.4 & 14.9 &  & 8.4 & 7.9 \\
    ToMe merge$^{\ast}$ &  &  &  &  &  & 18.8 &  &  \\
    ToMe RoPE$^{\ast}$ &  &  &  &  &  & 11.5 &  &  \\
    Blocks and rest &  & 56.2 & 51.9 & 45.3 & 34.0 & 72.3 & 67.8 & 60.5 \\
    \midrule
    Backbone, total & 91.9 & 69.1 & 65.6 & 59.7 & 48.9 & 102.6 & 76.2 & 68.4 \\
    Neck & 9.3 & 9.4 & 9.5 & 9.6 & 9.7 & 9.7 & 9.3 & 8.5 \\
    Head, total & 78.0 & 78.2 & 78.3 & 78.4 & 78.5 & 78.5 & 78.4 & 71.9 \\
    Head, encoder & 60.6 & 60.6 & 60.6 & 60.6 & 60.6 & 60.6 & 60.5 & 54.2 \\
    Head, decoder & 9.4 & 9.5 & 9.5 & 9.5 & 9.5 & 9.5 & 9.8 & 9.8 \\
    Head, other & 8.1 & 8.2 & 8.2 & 8.3 & 8.4 & 8.4 & 8.1 & 7.9 \\
    \midrule
    Sum of the three & 179.2 & 156.6 & 153.3 & 147.7 & 137.0 & 190.8 & 163.9 & 148.9 \\
    End to end, synchronised & 179.6 & 157.0 & 153.7 & 148.0 & 137.4 & 191.2 & 164.3 & 149.3 \\
    End to end, unsynchronised & 178.5 & 155.0 & 151.5 & 145.7 & 135.2 & 181.3 & 162.7 & 147.5 \\
    \bottomrule
  \end{tabular}
\end{table*}

\section{Attention kernel}
\label{sec:suppl2:kernel}

The block-diagonal variable-length call of \cref{sec:method} is one call per block to the xFormers memory-efficient attention kernel (xFormers \texttt{0.0.22.post7}, PyTorch \texttt{2.1.0}), with a bias built from the group lengths, so no group is padded. The code takes this path only when xFormers is installed and the activations are fp16 or bf16; otherwise the same grouping runs as masked attention in plain PyTorch, which pads every group to the longest one and calls \texttt{scaled\_dot\_product\_attention} with a key-padding mask. Every GFLOPs, latency and memory value in this paper is measured on the xFormers path.

\section{Lost objects}
\label{sec:suppl2:lost}

\begin{figure*}[t]
  \centering
  \includegraphics[width=\textwidth]{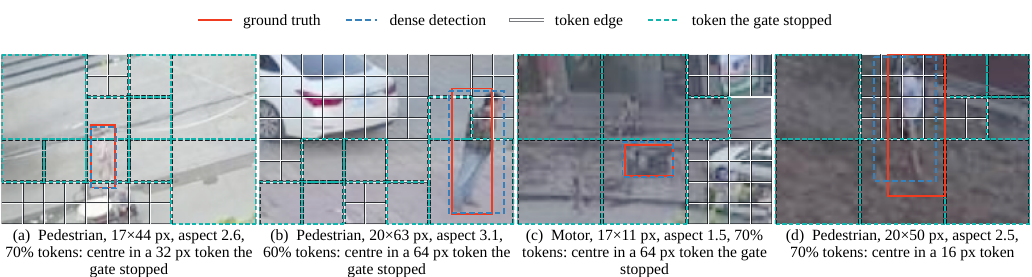}
  \caption{Four objects the dense detector finds and VGTok loses, at the budget given under each panel. In (a) to (c) the object's centre lies in a token the gate stopped; in (d), the more common case, it lies in a $16$-pixel patch. Ground truth is red, the dense detection dashed blue, and tokens the gate stopped dashed teal.}
  \label{fig:suppl2:lost}
\end{figure*}

\Cref{tab:suppl2:lost} compares VGTok at $70$ and $60\%$ of tokens with the dense detector on VisDrone val, with no retraining; an object is lost if the dense detector finds it at IoU $0.5$ and VGTok does not. A token the gate stopped is present in $\mathcal{T}+g$ and absent from $\mathcal{T}$ at the same threshold, so it is the coarse token of a node the gate left unsplit. \Cref{fig:suppl2:lost} shows four lost objects.

\begin{table*}
  \caption{Objects VGTok loses against the dense detector, with no retraining; VisDrone val, $548$ images, $38{,}759$ boxes, maxDets $500$. Recall is the share of boxes matched at IoU $0.5$. Second block: elongated (aspect ratio at least $2.5$) minus compact (below $1.5$), with its $95\%$ bootstrap interval over images; the loss rate is the lost objects over all objects of that shape. Third block: the token at the centre of each lost object, and the loss rate by that token, with the number of objects in brackets in its last four rows.}
  \label{tab:suppl2:lost}
  \centering
  \scriptsize
  \setlength{\tabcolsep}{4pt}
  \begin{tabular}{@{}lrrrr@{}}
    \toprule
    & Objects & Dense & 70\% & 60\% \\
    \multicolumn{5}{@{}l}{\emph{Recall at IoU $0.5$, by aspect ratio (long side over short side) and by size}} \\
    aspect $<1.5$ & 15{,}545 & 97.25 & 97.04 & 96.69 \\
    aspect $1.5$--$2.5$ & 16{,}653 & 96.93 & 96.56 & 96.14 \\
    aspect $2.5$--$4$ & 6{,}284 & 95.66 & 95.50 & 94.99 \\
    aspect $>4$ & 277 & 85.92 & 85.56 & 85.56 \\
    size small & 26{,}575 & 95.63 & 95.28 & 94.78 \\
    size medium & 11{,}116 & 99.22 & 99.14 & 98.93 \\
    size large & 1{,}068 & 99.81 & 99.53 & 99.53 \\
    \addlinespace[2pt]
    all objects & 38{,}759 & 96.77 & 96.50 & 96.10 \\
    \midrule
    \multicolumn{5}{@{}l}{\emph{Elongated (aspect $\geq 2.5$) minus compact (aspect $<1.5$), percentage points, with 95\% bootstrap interval over images}} \\
    net recall change &  &  & $+0.04$ [$-0.27$, $+0.35$] & $-0.09$ [$-0.46$, $+0.29$] \\
    loss rate &  &  & $+0.23$ [0.00, $+0.46$] & $+0.36$ [$+0.03$, $+0.68$] \\
    \midrule
    \multicolumn{5}{@{}l}{\emph{Token at the centre of each object lost against the dense detector}} \\
    objects lost &  &  & 224 & 409 \\
    centre in a $16$\,px token &  &  & 80.8\% & 64.5\% \\
    centre in a $32$\,px token &  &  & 9.8\% & 21.5\% \\
    centre in a $64$\,px token &  &  & 9.4\% & 13.9\% \\
    centre in a $32$ or $64$\,px token &  &  & 19.2\% & 35.5\% \\
    centre in a token the gate stopped &  &  & 18.8\% & 26.2\% \\
    loss rate, $16$\,px centre token &  &  & 0.47\% & 0.70\% \\
    loss rate, $32$\,px centre token &  &  & 7.36\% & 12.27\% \\
    loss rate, $64$\,px centre token &  &  & 30.88\% & 31.32\% \\
    loss rate, centre in a token the gate stopped &  &  & 13.38\% & 16.98\% \\
    \quad aspect $<1.5$ &  &  & 10.48\% (105) & 13.45\% (223) \\
    \quad aspect $1.5$--$2.5$ &  &  & 13.29\% (158) & 16.99\% (306) \\
    \quad aspect $2.5$--$4$ &  &  & 18.75\% (48) & 26.60\% (94) \\
    \quad aspect $>4$ &  &  & 33.33\% (3) & 0.00\% (7) \\
  \end{tabular}
\end{table*}

\section{Intervals}
\label{sec:suppl2:intervals}

\Cref{tab:suppl2:intervals} gives a paired bootstrap interval over images for each difference between VGTok and the dense detector at equal GFLOPs and at equal latency, and between VGTok and ToMe. Each of $2{,}000$ replicates draws image ids with replacement, every setting uses the same draw, and the interval is the $2.5$th to $97.5$th percentile of the differences. The dense AP at equal cost is interpolated linearly in cost between the two input sizes that bracket VGTok's cost, with weights fixed from the measured costs.

\begin{table}
  \caption{VGTok minus the dense detector at equal end-to-end GFLOPs or equal median latency, and minus ToMe, with no retraining; VisDrone val, $548$ images, and test-dev, $1{,}610$ images. Differences are in AP points on unrounded values, with their $95\%$ interval over $2{,}000$ paired replicates. Latencies on the packed attention are from the run that timed it. $^{\dagger}$Extrapolated below the smallest packed input size.}
  \label{tab:suppl2:intervals}
  \centering
  \scriptsize
  \setlength{\tabcolsep}{2pt}
  \begin{tabular}{@{}lrrrr@{}}
    \toprule
    & \multicolumn{2}{c}{AP} & \multicolumn{2}{c}{AP$_S$} \\
    \cmidrule(lr){2-3}\cmidrule(l){4-5}
    VGTok, matched on & Difference & $95\%$ interval & Difference & $95\%$ interval \\
    \midrule
    \multicolumn{5}{@{}l}{\emph{val, against dense on its default attention}} \\
    $80\%$, GFLOPs & $+0.54$ & [$+0.30$, $+0.75$] & $+0.73$ & [$+0.34$, $+1.12$] \\
    $80\%$, latency & $+0.54$ & [$+0.31$, $+0.76$] & $+0.74$ & [$+0.35$, $+1.13$] \\
    $70\%$, GFLOPs & $+0.62$ & [$+0.32$, $+0.88$] & $+1.14$ & [$+0.68$, $+1.59$] \\
    $70\%$, latency & $+0.37$ & [$+0.11$, $+0.60$] & $+0.73$ & [$+0.31$, $+1.11$] \\
    $60\%$, GFLOPs & $+0.30$ & [$-0.08$, $+0.62$] & $+0.94$ & [$+0.48$, $+1.37$] \\
    $60\%$, latency & $-0.35$ & [$-0.72$, $-0.03$] & $+0.23$ & [$-0.27$, $+0.72$] \\
    $40\%$, GFLOPs & $-5.16$ & [$-5.82$, $-4.55$] & $-3.74$ & [$-4.57$, $-2.91$] \\
    $40\%$, latency & $-6.12$ & [$-6.78$, $-5.52$] & $-4.94$ & [$-5.73$, $-4.16$] \\
    \multicolumn{5}{@{}l}{\emph{val, against dense on the packed attention VGTok uses}} \\
    $80\%$, GFLOPs & $+0.45$ & [$+0.20$, $+0.67$] & $+0.59$ & [$+0.17$, $+0.98$] \\
    $80\%$, latency & $+0.08$ & [$-0.05$, $+0.21$] & $+0.12$ & [$-0.08$, $+0.33$] \\
    $70\%$, GFLOPs & $+0.52$ & [$+0.24$, $+0.76$] & $+0.99$ & [$+0.54$, $+1.39$] \\
    $70\%$, latency & $-0.11$ & [$-0.31$, $+0.08$] & $+0.03$ & [$-0.31$, $+0.31$] \\
    $60\%$, GFLOPs & $+0.05$ & [$-0.33$, $+0.34$] & $+0.70$ & [$+0.21$, $+1.11$] \\
    $60\%$, latency & $-0.85$ & [$-1.21$, $-0.54$] & $-0.61$ & [$-1.19$, $-0.11$] \\
    \multicolumn{5}{@{}l}{\emph{val, against ToMe}} \\
    $80\%$ minus $m=10$ & $+2.84$ & [$+2.58$, $+3.12$] & $+3.97$ & [$+3.55$, $+4.38$] \\
    \midrule
    \multicolumn{5}{@{}l}{\emph{test-dev, against dense on its default attention}} \\
    $80\%$, GFLOPs & $+0.36$ & [$+0.14$, $+0.54$] & $+0.57$ & [$+0.26$, $+0.86$] \\
    $80\%$, latency & $+0.30$ & [$+0.08$, $+0.48$] & $+0.48$ & [$+0.18$, $+0.78$] \\
    $70\%$, GFLOPs & $+0.13$ & [$-0.11$, $+0.37$] & $+0.57$ & [$+0.20$, $+0.89$] \\
    $70\%$, latency & $-0.06$ & [$-0.30$, $+0.15$] & $+0.28$ & [$-0.08$, $+0.60$] \\
    $60\%$, GFLOPs & $-0.36$ & [$-0.65$, $-0.05$] & $+0.55$ & [$+0.14$, $+0.91$] \\
    $60\%$, latency & $-0.71$ & [$-1.00$, $-0.45$] & $-0.02$ & [$-0.42$, $+0.34$] \\
    $40\%$, GFLOPs & $-4.72$ & [$-5.24$, $-4.22$] & $-2.06$ & [$-2.63$, $-1.52$] \\
    $40\%$, latency & $-5.48$ & [$-6.00$, $-4.97$] & $-3.00$ & [$-3.61$, $-2.50$] \\
    \multicolumn{5}{@{}l}{\emph{test-dev, against dense on the packed attention VGTok uses}} \\
    $80\%$, GFLOPs & $+0.30$ & [$+0.08$, $+0.48$] & $+0.44$ & [$+0.16$, $+0.74$] \\
    $80\%$, latency & $-0.03$ & [$-0.18$, $+0.09$] & $+0.02$ & [$-0.14$, $+0.18$] \\
    $70\%$, GFLOPs & $+0.05$ & [$-0.19$, $+0.27$] & $+0.41$ & [$+0.06$, $+0.74$] \\
    $70\%$, latency & $-0.43$ & [$-0.63$, $-0.25$] & $-0.25$ & [$-0.52$, $-0.02$] \\
    $60\%$, GFLOPs$^{\dagger}$ & $-0.51$ & [$-0.83$, $-0.20$] & $+0.27$ & [$-0.19$, $+0.71$] \\
    $60\%$, latency & $-1.10$ & [$-1.37$, $-0.85$] & $-0.60$ & [$-0.95$, $-0.25$] \\
    \bottomrule
  \end{tabular}
\end{table}

\end{document}